\pdfoutput=1

\documentclass[final]{cvpr}

\usepackage{times}
\usepackage{epsfig}
\usepackage{graphicx}
\usepackage{amsmath}
\usepackage{amssymb}
\usepackage{tabularx}
\usepackage{multirow}
\usepackage{xcolor}
\usepackage{ulem}
\usepackage{colortbl}
\usepackage[]{algorithm2e}
\usepackage{mathtools}

\newcommand{\comment}[1]{}
\newcommand{\parag}[1]{\vspace{-3mm}\paragraph{#1}}

 \newcommand{\PF}[1]{{\color{red}{\bf PF: #1}}}

 \newcommand{\nd}[1]{{\color{orange} #1}}
 
 \usepackage{amsmath,amsfonts,bm}

\def\eqref#1{equation~\ref{#1}}
\def\1{\bm{1}}

\def\vtheta{{\bm{\theta}}}

\def\vb{{\bm{b}}}

\def\vm{{\bm{m}}}

\def\vx{{\bm{x}}}
\def\vy{{\bm{y}}}
\def\vz{{\bm{z}}}

\def\vN{{\textbf{N}}}
\def\vM{{\textbf{M}}}
\def\vS{{\textbf{S}}}

\def\vW{{\textbf{W}}}

\DeclareMathAlphabet{\mathsfit}{\encodingdefault}{\sfdefault}{m}{sl}
\SetMathAlphabet{\mathsfit}{bold}{\encodingdefault}{\sfdefault}{bx}{n}

\def\sR{{\mathbb{R}}}

\usepackage[pagebackref=true,breaklinks=true,colorlinks,bookmarks=false]{hyperref}

\def\cvprPaperID{7469} % *** Enter the CVPR Paper ID here

\def\cvprPaperID{7469} % *** Enter the CVPR Paper ID here
\def\confYear{CVPR 2021}

\begin{document}

\title{Masksembles for Uncertainty Estimation}  % **** Enter the paper title here

\author{
    Nikita Durasov\textsuperscript{1}\thanks{This work was supported in part by the Swiss National Science Foundation }
    \quad
    Timur Bagautdinov\textsuperscript{2}
    \quad
    Pierre Baque\textsuperscript{2}
    \quad
    Pascal Fua\textsuperscript{1}\\
    \textsuperscript{1}Computer Vision Laboratory, EPFL, {\tt\small \{name.surname\}@epfl.ch}\\
    \textsuperscript{2}Neural Concept SA, {\tt\small \{name.surname\}@neuralconcept.com} \vspace{2.5mm} \\
    {\small \url{https://nikitadurasov.github.io/projects/masksembles/}}\\\
    % \textsuperscript{3}Facebook Reality Labs Research, {\tt\small timurb@fb.com}\\
}
\pagenumbering{gobble}
\maketitle

% \author{Durasov Nikita\\
% CVLAB, EPFL\\
% {\tt\small nikita.durasov@epfl.ch}

% \and

% Timur Bagautdinov\\
% Facebook Research\\
% {\tt\small timurb@fb.com}

% \and

% Pierre Baque\\
% Neural Concept SA\\
% {\tt\small  pierre.baque@neuralconcept.com}

% \and

% Pascal Fua\\
% CVLAB, EPFL\\
% {\tt\small pascal.fua@epfl.ch}
% }

%%%%%%%%% BODY TEXT - ENTER YOUR RESPONSE BELOW
\begin{abstract}

Deep neural networks have amply demonstrated their prowess but estimating the reliability of their predictions remains challenging. 
Deep Ensembles are widely considered as being one of the
best methods for generating uncertainty 
estimates but are very expensive to train and evaluate. 

MC-Dropout is another popular alternative, which is less expensive, 
but also less reliable.
Our central intuition is that there is a continuous spectrum of ensemble-like 
models of which MC-Dropout and Deep Ensembles are extreme examples. 
The first one uses an effectively infinite number of highly correlated models, while the second relies on a finite number of independent models. 

To combine the benefits of both, we introduce {\it Masksembles}. 
Instead of randomly dropping parts of the network as in MC-dropout, 
Masksemble relies on a fixed number of binary masks, which are
parameterized in a way that allows changing correlations between 
individual models. Namely, by controlling the overlap between the 
masks and their size one can choose the optimal
configuration for the task at hand.
This leads to a simple and easy to implement method with
performance on par with Ensembles at a fraction of the cost. 
We experimentally validate Masksembles on two widely used datasets, CIFAR10 and ImageNet.

%----
% OLD
%----

% with a specified number of ones and a parametrizable overlap between them. 
% This adds structure to the activation zeroing-out and allows us 
% to find optimal configurations.

%Based on that, we propose Masksembles, a method that combines  the benefits of both approaches, while making it possible to control the trade-off between model performance and computational costs.
%
% We experimentally validate our method on  and demonstrate that one can achieve close to state-of-the-art performance at a fraction of computational cost.
%
%Our study also provides experimental and theoretical insights into MC-Dropout and Ensembles, as those can be represented as limit cases of Masksembles.

% In practice, it amounts to introducing fixed set of binary masks
% which define which network parameters should be dropped. 

% In practice, we the key idea is to introduce a discrete set of binary masks which define which network weights should be dropped and which should be 
% used. By controlling the correlation between these masks, we can obtain Ensemble-like uncertainty estimates at a computational cost comparable to that of MC-Dropout. Since the parameters can be tuned to reproduce Ensembles and Dropout as limit cases of Masksembles, the present study offers experimental and theoretical insights into both approaches.
\end{abstract}
\section{Introduction}

The ability of deep neural networks to produce useful predictions is now
abundantly clear~\cite{vaswani2017attention, chen2017deeplab, redmon2016you, silver2016mastering, dosovitskiy2020image, durasov2019double, ivanov2018learning, kruzhilov2019double} but assessing the reliability of these predictions remains a
challenge. 
Among all the methods that can be used to this end, MC-Dropout~\cite{Gal16} and 
Deep Ensembles~\cite{Lakshminarayanan17} have emerged as two of the most popular 
ones.  
Both of those methods exploit the concept of \textit{ensembles} 
to produce uncertainty estimates. 
%
% Namely, these methods produce
% a diversified collection of models, and then simultaneously 
% apply those models to the given inputs to produce multiple 
% versions of predictions, which are then aggregated to obtain 
% an uncertainty measure. 
%
MC-Dropout does so implicitly - by training a single \textit{stochastic}
network~\cite{ghiasi2018dropblock, wan2013regularization, kingma2015variational}, where randomness is achieved by dropping different subsets of weights 
for each observed sample. At test time, one can obtain multiple
predictions by running this network multiple times, each time
with a different weight configuration.
Deep Ensembles, on the other hand, build an explicit ensemble of models,
where each model is randomly initialized and trained independently 
using \textit{stochastic} gradient descent~\cite{ruder2016overview, kingma2014adam}. 
As importantly, both methods rely on the fact that the outputs of individual 
models are diverse, which is achieved by introducing stochasticity into 
the training or testing process.

However, simply adding randomness does not always lead to diverse predictions. 
In practice, MC-Dropout often performs significantly worse than Deep Ensembles 
on uncertainty estimation tasks~\cite{Lakshminarayanan17, ovadia2019can, gustafsson2020evaluating}.  
We argue that this can be attributed to the fact that each weight is dropped 
randomly and independently from the others. For many configurations of 
hyperparameters, in particular the dropout rate, 
this results in similar weight configurations and, consequently, less diverse 
predictions~\cite{fort2019deep}.
Deep Ensembles seem to not share this weakness for reasons that are not yet fully understood and usually produce significantly more reliable uncertainty estimates. 
Unfortunately, this comes at a price, both at training and inference time: 
Building an ensemble requires training multiple models, and using it means 
loading all of them simultaneously in memory, 
typically a very scarce resource.

In this work, we introduce \textit{Masksembles}, an approach to uncertainty
estimation that tackles these challenges and produces reliable uncertainty 
estimates on par with Deep Ensembles at a significantly lower computational cost.
The main idea behind the method is simple - to introduce a more structured 
way to drop model parameters than that of MC-Dropout.

Masksembles produces a \textit{fixed number} of binary masks which
specify the network parameters to be dropped. 
The properties of the masks effectively define the properties of the final ensemble in terms of capacity and correlations.
%and \nd{and increases the number of activations used in layers}. 
%
During training, for each sample, we randomly choose one of the masks and drop the corresponding parts of the model just like a standard dropout. %
During inference, we run the model multiple times, once per mask,
to obtain a set of predictions and, ultimately, an uncertainty estimate.
Our method has three key hyperparameters: the total number of masks, the 
average overlap between masks and the number of ones and zeros 
in each mask.
Intuitively, using a large number of masks approximates MC-Dropout, while 
using a set of non-overlapping, completely disjoint masks yields an 
Ensemble-like behavior.
In other words, as illustrated in Fig.~\ref{fig: mcdp-ensembles-transition} Masksembles defines a spectrum of model configurations 
of which MC-Dropout and Deep Ensembles can be seen as extreme cases. 
We evaluate our method on several synthetic and real datasets and 
demonstrate that it outperforms MC-Dropout in terms of accuracy,
calibration, and out-of-distribution (OOD) detection performance.
When compared to Deep Ensembles, our method has similar performance but is more favorable in terms
of training time and memory requirements.
Furthermore, Masksembles is simple to implement and can serve as a drop-in replacement for MC-Dropout in virtually any model. 
To summarize, the main contributions of this work are as follows:
\begin{itemize}
    \item We propose an easy to implement, drop-in replacement method that performs better than MC-Dropout, at a similar computational cost, and 
    that matches Ensembles at a fraction of the cost.
    \item We provide theoretical insight into two popular 
    uncertainty estimation methods - MC-Dropout and Ensembles,
    by creating a continuum between the two.
    \item We provide a comprehensive evaluation of our method on 
    several public datasets. This validates our claims and 
    guides the choice of Masksemble parameters
    in practical applications. 
\end{itemize}

\section{Related Work}

MC-Dropout~\cite{Gal16} and Deep Ensembles~\cite{Lakshminarayanan17} have emerged as two of the most prominent and practical 
uncertainty estimation methods for deep neural networks~\cite{ashukha2020pitfalls, ovadia2019can, gustafsson2020evaluating}.
In what follows, we provide a brief background on uncertainty estimation, review the best-known methods, and discuss potential alternatives~\cite{wortsman2020supermasks, havasi2020training, atanov2019uncertainty, malinin2018predictive, malinin2020regression}.

\subsection{Background}

The goal of \textit{Uncertainty Estimation} (UE) is to produce a measure of confidence for model predictions.
Two major types of uncertainty can be 
modeled within the Bayesian framework~\cite{der2009aleatory}.
\textit{Aleatoric uncertainty}, also known as \textit{Data 
uncertainty}~\cite{gal2016uncertainty}, captures noise that arises from data and therefore is irreducible, because the natural complexity of the data directly causes it.
Sensor noise, labels noise, and classes overlapping are among the most common sources of aleatoric uncertainty.
\textit{Epistemic uncertainty} or \textit{model 
uncertainty}~\cite{gal2016uncertainty} accounts for uncertainty in
the parameters of the model that we learned, that is, it 
represents our ignorance about parameters of the data generation 
process. 
This type of uncertainty is reducible and therefore, given enough 
data, could be eliminated entirely. 
In addition to the aforementioned types of uncertainty,
there is also \textit{Distributional uncertainty}. 
Also known as \textit{dataset shift}~\cite{quionero2009dataset},
this type of uncertainty is useful when there is a mismatch
between the training and testing data distributions, and a model 
is confronted with unfamiliar samples.

For all the aforementioned types of uncertainty, ideally we
want our model to output an additional signal indicating
how reliable its prediction is for a particular sample. 
More formally, given an input, we want our model to output the 
actual prediction target as well as the corresponding uncertainty
measure.
For regression tasks, the uncertainty measure can be
the variance or confidence intervals around the prediction 
made by the network~\cite{pearce2018high}.
For classification tasks, popular uncertainty measures
are the entropy and max-probability~\cite{malinin2018predictive, lakshminarayanan2017simple}.
Ideally, the measure of uncertainty should be high when the model
is incapable of producing accurate prediction for the given input 
and low otherwise. 
Generally speaking, uncertainty can be either learned from the data 
directly~\cite{lakshminarayanan2017simple, kendall2017uncertainties}, in particular for big 
data regimes and aleatoric uncertainty, 
or could be acquired from a 
diverse set of predictions~\cite{Gal16, lakshminarayanan2017simple},
which are obtained from stochastic regularization
techniques or ensembles. In this work, we focus on the latter family of methods.

\subsection{Ensembles}

Deep Ensembles~\cite{Lakshminarayanan17} involve training an
ensemble of deep neural networks on the same data by initializing
each network randomly. This approach is originally inspired by a 
classical ensembling method, bagging~\cite{Hastie09}.
An uncertainty estimate is obtained by running all the elements in the ensemble and aggregating their predictions.
The major drawback of Deep Ensembles is the 
computational overhead: it requires training 
multiple independent networks, and during inference 
it is desirable to keep all these networks in memory. 
More generally, ensembling has been a popular way to boost 
the performance of individual neural network models
~\cite{hansen1990NNensembles,krizhevsky2009learning,lakshminarayanan2017simple,perrone1992networks, yeo2021robustness, zamir2020robust}, with the quality improvements
often being associated with diverse predictions.
In~\cite{fort2019deep} authors provide a comprehensive comparison of several popular ensemble learning methods as well as
several approximate Bayesian inference techniques. 
Extensive experiments on vision tasks suggest 
that simple ensembles have the lowest correlation 
between individual models.
This is an important advantage of these methods, as 
ultimately this leads to highly diversified predictions
on the samples which are very distinct from those
observed in training.

Snapshot-based methods take a slightly different approach towards
creating an ensemble, and try to collect a diverse set
of weight configurations over the course of a single 
training.
Snapshot ensembles~\cite{huang2017snapshot} build upon
the idea of using cyclical learning rates to ensure that the
optimization process can efficiently explore multiple 
local optima.
Similarly, Fast Geometric Ensembling~\cite{Garipov18, izmailov2018averaging},
uses a method to identify low-error paths between local
optima to obtain a collection of diverse representations.
Although these approaches partially tackle training time issues,
they typically perform worse than the standard ensembles~\cite{ashukha2020pitfalls} and consume the
same amount of memory. 
Furthermore, because of the specifics of their training
procedure, in particular, shared initialization with
a pre-trained network, individual snapshots are not guaranteed
to be decorrelated, which might hurt the performance~\cite{ashukha2020pitfalls}.

\subsection{Dropout}

Dropout~\cite{Srivastava14} is an extremely popular and simple
approach to stochastically regularizing deep neural networks. 
%
% Even though its performance has often been question, the simplicity and ease of use of the approach makes it very widely implemented.
During training, neurons are randomly dropped with fixed 
probability, which effectively produces an approximation of 
several slightly different model architectures and
allows a single model to mimic ensemble behavior.
Originally Dropout was used during training as a stochastic
regularization technique, but~\cite{Gal16, gal2016uncertainty} 
proposed using it during inference, providing sound mathematical 
justification for ensembling and Bayesian model averaging analogies.
MC-Dropout is extremely easy to use and has zero memory 
overhead compared to a single model but shows consistently
worse performance compared to Deep Ensembles~\cite{ovadia2019can, gustafsson2020evaluating}. 
In particular, different predictions made by several forward passes with randomly generated masks seem to be overly correlated and strongly underestimate the variance.
In this paper, we postulate that the training procedure, which makes any combination of the weights possible, will tend to create uniformity between predictions. 
As opposed to Ensemble techniques, MC-Dropout techniques do not let
multiple versions of the network actually be created, and the
weights are forced to converge jointly.

\subsection{Other Methods}

Although Bayesian inference provides a natural way to assess 
uncertainty of model predictions, it is prohibitively expensive to apply these techniques to deep neural networks directly.
A number of approximate inference techniques were developed for \textit{Bayesian Neural Networks} (BNN)~\cite{mackay1992bayesian, neal2012bayesian, mackay1992practical} that try to alleviate these problems, to name a few: \textit{Bayes by Backprop}~\cite{blundell2015weight} uses
Variational Inference to learn a parametric distribution over model weights that approximates the true posterior, 
\textit{Laplace Approximation}~\cite{ritter2018scalable} tries to 
find a Gaussian approximation to the posterior modes,
MC-Dropout~\cite{Gal16} could be seen as an approximate Bayesian 
procedure as well. 
Although these techniques are theoretically grounded,
 Deep Ensembles often show significantly better performance in practice~\cite{ovadia2019can, ashukha2020pitfalls},
in terms of both accuracy and quality of the resulting 
uncertainty estimates.

% , along with Bayesian 
% networks~\cite{mackay1992bayesian}, which are somewhat 
% less scalable \TB{Are they?}.
% MC-Dropout estimates the variance of predictions by running a 
% network multiple times, each time randomly dropping, or zeroing out, 
% some of the network weights. 
% %
% Deep Ensembles are essentially a deep version of a 
% classical method known as bagging. In a nutshell,
% it involves several independent models starting
% from random initializations. 
%
% In practice, Ensembles tend to perform better but at the cost of 
% being far more computationally demanding. 

\section{Method}
\label{sec:method}

In this section, we introduce Masksembles, an approach to 
uncertainty estimation that creates a continuum between single 
networks, deep ensembles, and MC-Dropout. 
It builds on the intuition  that dropout-based methods can be
seen as stochastic variants of ensembles~\cite{lakshminarayanan2017simple}, albeit ensembles 
with infinite cardinality.
We will argue that the main reason for MC-Dropout's poor performance is the high correlation between the ensemble elements that makes the overall predictions insufficiently diverse. 
Masksembles is designed to give control over the correlation 
between the elements of an ensemble and, hence, achieve a  
satisfactory compromise between reliable uncertainty estimation 
and acceptable computational cost.

\subsection{Masksembles as Structured Dropout} 

Both Ensembles and MC-Dropout are ensemble methods, that is, they produce multiple predictions for a given input and then
use an aggregated measure such as variance or entropy
as an uncertainty estimator.
Formally, consider a dataset $\{ \vx_i, y_i \}_{i}$, where $\vx_i \in \sR^{F}$ are input features, and $y_i$ are
scalars or labels.
For classification problems, we model conditional distribution $p(y | \vx ; \vtheta)$ as 
a mixture
\begin{equation*}
    p(y | \vx ; \vtheta) = \frac{1}{N} \sum_{k=1}^{N} p(y | \vx ; \vtheta_{k}) \; ,
\end{equation*}
where $\vtheta_k$ are the weights of element $k$ of the
ensemble, and $N$ is the size of the ensemble.

For Deep Ensembles, individual models are independent and do not share any weights, and hence each $\vtheta_k$ is a separate 
set of weights, trained entirely independently.  
By contrast, for MC-Dropout there is a single shared
$\vtheta$, and individual models  
are obtained as $\vtheta_k = \tilde{\vb}^t_k \cdot \vtheta$,
where $\tilde{\vb}^t_k \in \{ 0, 1 \}^{|\vtheta|}$ 
are random binary masks which are re-sampled for each
iteration $t$. 
The elements of $\tilde{\vb}^t_k$ follow Bernoulli 
distribution parameterized by dropout probability $p$, 
and the parameters corresponding to the same network activations are dropped simultaneously.
At inference time, a small number of predictions is produced 
by randomly sampling the masks. In practice, 
for any reasonable choice of $p$, the masks $\tilde{\vb}^t_k$ will 
overlap significantly, which will tend to make predictions 
$p(y | \vx ; \vtheta_k)$ highly correlated, and thus could lead to 
underestimated uncertainty. 
Furthermore, because masks are re-sampled at each training 
iteration, each network unit needs to be able to form a 
coherent response with any other unit, which creates
a mixing effect and leads to uniformity between 
the predictions from different masks.
% % 
% This may be one of the reasons MC-dropout often underestimates uncertainty.

To tackle these issues, we propose using a finite set of pre-defined binary masks, which are generated so that their
overlap can be controlled. 
They are then used to drop corresponding network activations
so that the resulting models are sufficiently decorrelated and do not suffer from the mixing effect described above. 
In what follows, we describe the algorithm for mask generation, 
explain the way we apply generated masks to models, and discuss the trade-offs it incurs.

\subsection{Generating Masks} 
\label{sec:method:generating-masks}

\begin{figure}
\includegraphics[width=0.49\textwidth]{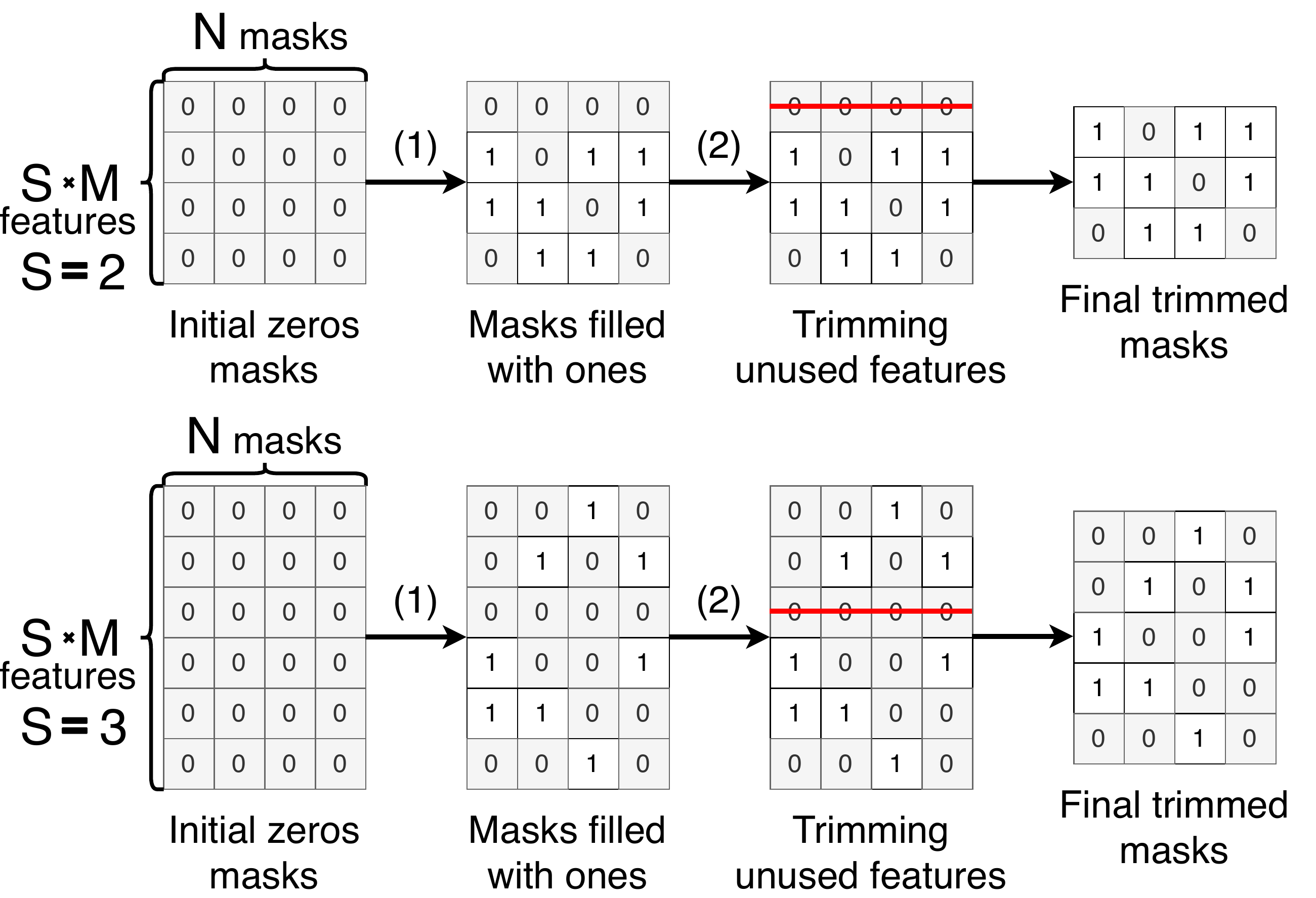}
\centering
\caption{\small {\bf Masks generation algorithm.} Here we use $\vN = 4$, $\vM = 2$, $\vS = 2.0 \text{ and } 3.0$. In the first case $\vM \cdot \vS = 4$ and in the second $\vM \cdot \vS = 6$. First, we generate 4 masks with $\vM \cdot \vS$ zeros each. Second, we randomly choose 2 positions for every of 4 masks and fill them with ones. Finally, we drop features that are not used in any mask. In the first case, the IoU is 0.44 and in the second 0.16.}
\label{fig: dropping features}
\end{figure}

Generating masks with a controllable amount of overlap 
is central to Masksembles. 
Let us first introduce the key parameters of our mask
generation process:
\begin{itemize}
\setlength\itemsep{0mm}
    \item $\vN$ - number of masks. %Size of pre-generated set of masks  $\{\tilde{\vm}_i\}_{i=1}^N$ that can be randomly picked at each step of training or inference.
    
    \item $\vM$ - number of ones in each mask. %Setting $\vM$ to particular value will guarantee that after masksare generated each of them includes exactly $\vM$ ones.
    
    \item $\vS$ - scale that controls the amount of overlap given $\vN$ and $\vM$. 
    %. The role of this parameter will be explained in section~\ref{sec:method:generating-masks}. To build intuition it is worth mentioning that, in the dropout limit $\vN \xleftarrow \infty$, then $\vS = 1 - \frac{1}{p}$, where $p$ is the dropout rate.
    
\end{itemize}
%
% The mask generation algorithm is summarized in 
% Alg.~\ref{alg: masks generation}.
The masks generation algorithm is summarized below. It starts by generating $\vN$ vectors of all zeros of 
size $\vM \times \vS$.
Then, in each of those vectors, it randomly sets $\vM$ of these elements to be 1 in each vector. 
As depicted by Fig.~\ref{fig: dropping features}, it then looks for positions that are all zero in the resulting vectors. These correspond to features that will be dropped.

For $\vS = 1.0$, the algorithm then generates $\vN$ masks of size $\vM$ that will contain only ones and therefore fully overlapping. 
Conversely, setting $\vS$ to large positive values will generate $\vN$ masks with mostly zero values and therefore very few overlapping one values.

% \begin{algorithm}[hThe mask generation algorithm is summarized in 
% Alg.~\ref{alg: masks generation}!]
%  \KwIn{$\vN, \vM, \vS$}
%  \KwResult{set of $N$ masks with defined properties}
%  \textit{masks} = [$\vN$ zeros vectors with $\vM \cdot \vS$ elements each]\;
%  \For{\textit{mask} in \textit{masks}} {
%  pick $\vM$ positions in $\textit{mask}$ (without replacement)\;
%  fill picked positions with ones\; 
%  }
%  drop unused features from \textit{masks}\;
%  \Return \textit{masks}
% \label{alg: masks generation}
% \vspace*{2mm}
% \caption{Procedure to generate set of binary masks with controllable average overlapping value.}
% \end{algorithm}

\subsection{Inserting Masksembles Layer into Networks}
\label{sec:insterting}

\begin{figure}[h!]
\includegraphics[width=0.49\textwidth]{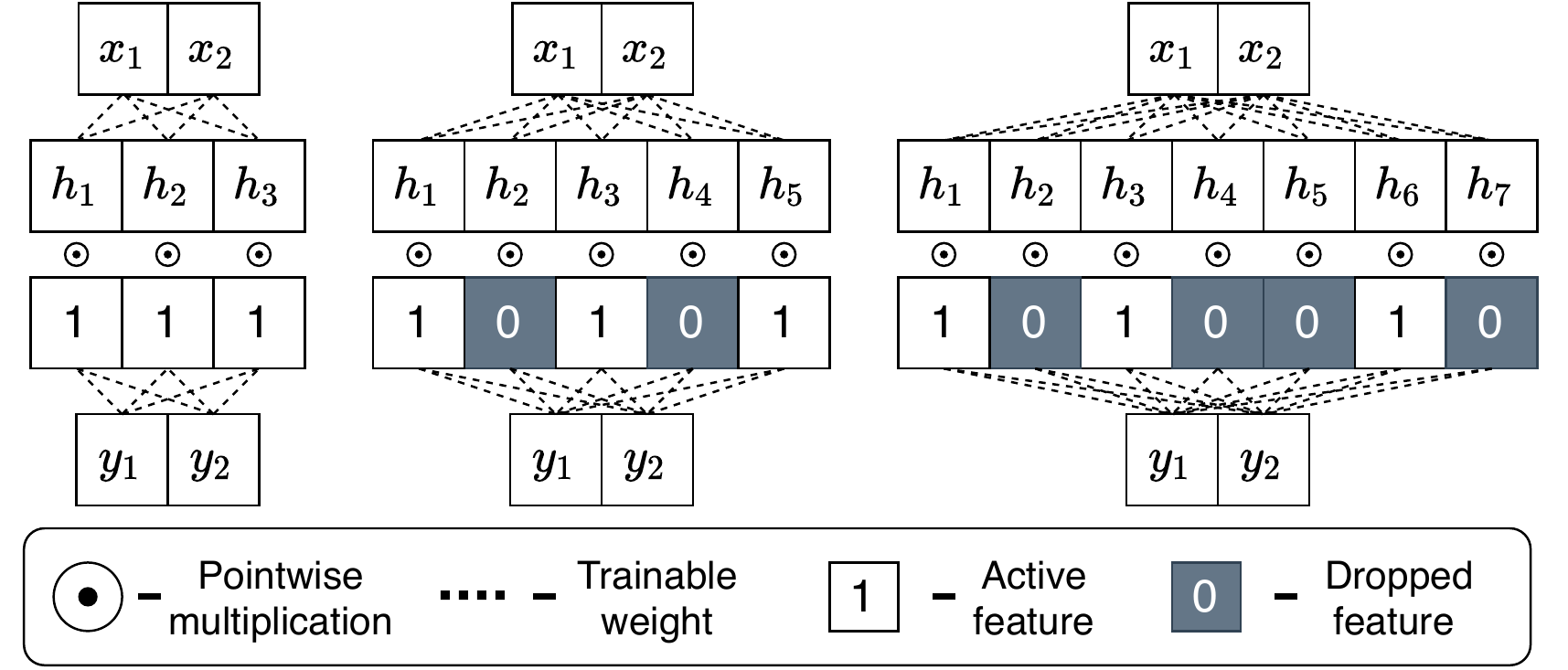}
\centering
\caption{\small{\bf Masksembles in a simple case.} Example of Masksembles MLP behavior for different 
values of $\vS \in [1.0, 1.7, 2.3]$ ($\vN$ and $\vM$ are fixed): $x_k$ - inputs, $h_k$ - activations, $y_k$ - outputs.}
\label{fig:masksembles-mlp}
\end{figure}

We start with the very simple case depicted by Fig.~\ref{fig:masksembles-mlp}. It shows an MLP with one hidden layer of size 3 and 2-dimensional inputs and outputs. We insert Masksembles layer right after the hidden layer and increase its size to $\vM \times \vS - D$ where $D$ is the number of dropped features so that the sizes of the masks and the size of the layer match. Then at run-time, we apply the mask values and ignore all components of the hidden layer for which they are zero.

Fig.~\ref{fig:masksembles-mlp} features 3 different configurations obtained by fixing both $\vM$ and $\vN$ --- here, $\vN$ is sufficiently large so that $D=0$ --- and changing the value of $\vS$ to 1.0, 1.7, and 2.3, therefore increasing masks sizes and hidden size in range [3, 5, 7].

Applying such modification to the model creates a larger one that allows for multiple relatively uncorrelated predictions. This process extends naturally to MLPs of any dimension. It also extends to convolutional architectures without any significant modification. As in MC-dropout, we can drop entire channels~\cite{Tompson15} instead of individual activations. In other words, treating the whole channel as a single feature
is equivalent to the fully connected case discussed above. 

In general, this means that we may have to adapt the size of network layers each time we change $\vM$, $\vN$, or $\vS$, which can be cumbersome. Fortunately, keeping $\vS \times \vM$ equal to the original channels number,  3 in the MLP example, makes it possible to apply Masksembles without any such change of the network configuration.

\subsection{Parameters and Model Transitions}
\label{sec:transition}

\begin{figure}[h]
\centering
\hspace*{-0.3cm}\begin{tabular}{@{}c@{}c@{}}
  \includegraphics[width=0.3\textwidth]{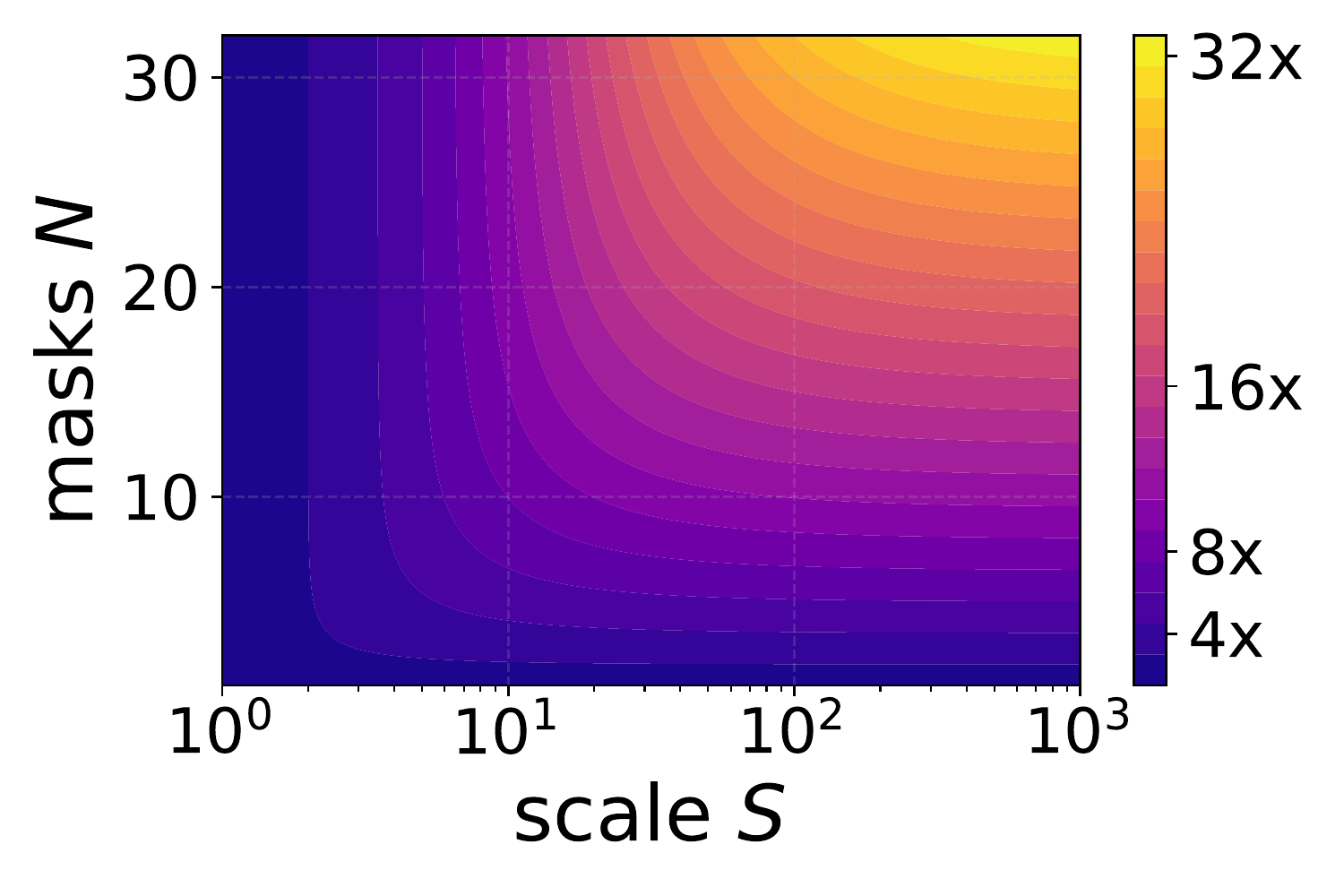} &   \includegraphics[width=0.2\textwidth]{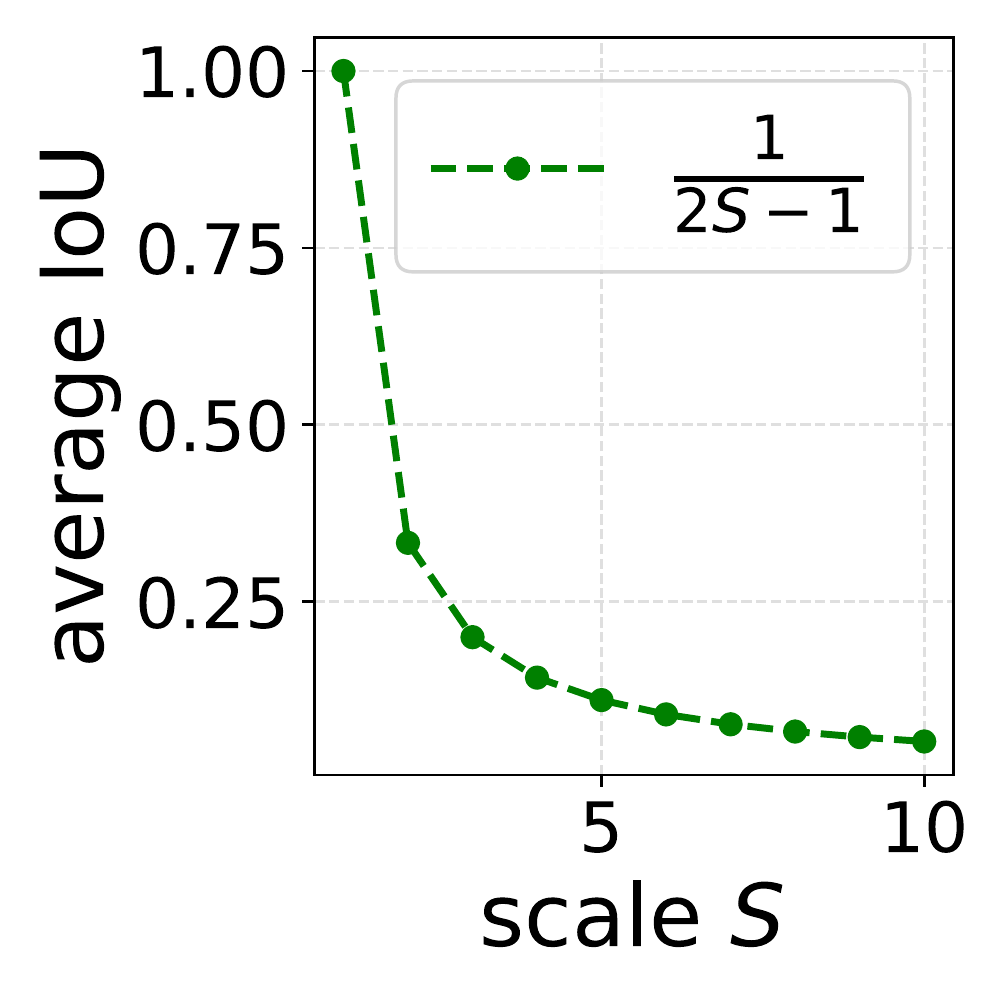} \\
\end{tabular}
\caption{\small {\bf Influence of the parameters.} \textbf{(Left)} Relative \textit{Total model size} as a function of $\vS$ and $\vN$, with $\vM$ being fixed. We refer to size of single model ($\vS = 1, \vN = 1$) as 1x. \textbf{(Right)} IoU of the generated masks as a function of $\vS$. 
Note that it does not depend on neither $\vN$ nor $\vM$.}
\label{fig: IoU Vs S}
\end{figure}

The chosen parameters $\vN$, $\vM$ and $\vS$ define several key model properties:
\begin{itemize}
\setlength\itemsep{0mm}

    \item \textit{Total model size}: The total number of model parameters. \comment{As illustrated in Fig.~\ref{fig: dropping features}, ~\ref{fig:masksembles-mlp}, it is equal to $\vM \times \vS - D$ where $D$ is the number of dropped features, which can be computed after the masks have been populated.} In the above MLP example, this is equal to $4 \cdot \left(\vM \times \vS - D\right)$ where $D$ is the number of features trimmed during the mask generation procedure. This functional relationship is illustrated in Fig.~\ref{fig: IoU Vs S}.
    
    \item \textit{Model capacity}: The number of activations that are used during one forward pass of the model, which is equal to  $\vM$.
    
    \item \textit{Masks IoU} average \textit{ Intersection over Union (IoU)} between generated masks. We derive approximation for this quantity that appears to depend only on $\vS$ : $\mathrm{IoU}(\vS) = 1 / (2 \vS - 1)$ and therefore $\vN, \vM$ do not influence it, as show in Fig.~\ref{fig: IoU Vs S}.
    
\end{itemize}

\begin{figure}
\hspace*{-0.4cm}
\includegraphics[width=0.49\textwidth]{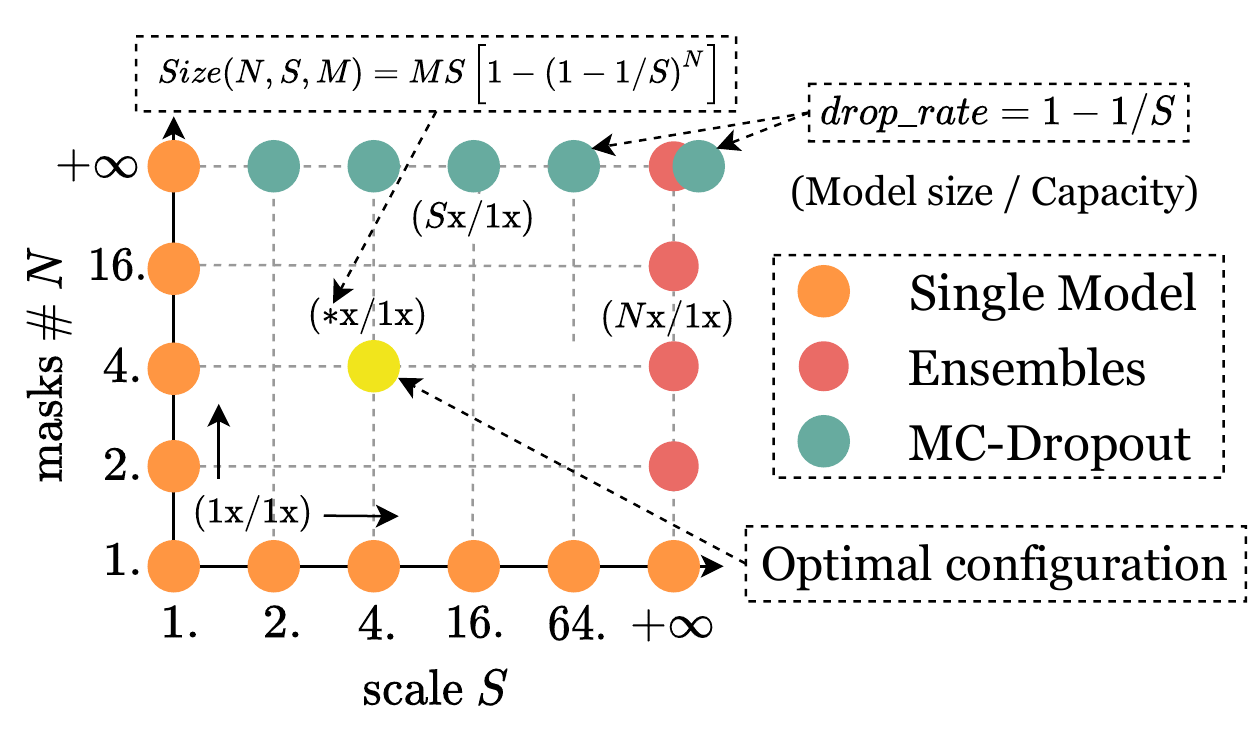}
\centering
% \vspace*{-0.2cm}
\caption{\small {\bf Ensembles to MC-Dropout transition}. \textcolor[HTML]{67AB9F}{\textbf{Green}} / \textcolor[HTML]{EA6B66}{\textbf{Red}}
 points denote extreme cases that correspond to these models, whereas the \textcolor[HTML]{E3C800}{\textbf{Yellow}} point represents an optimal parameter choice one could choose for a specific computational cost vs performance trade-off. Here $\vM$ is fixed while $\vN$ and $\vS$ vary. }
\label{fig: mcdp-ensembles-transition}
\end{figure}

In fact, Deep Ensembles and MC-Dropout can be both seen as extreme cases and Masksembles can transition smoothly from one to the other (Fig.~\ref{fig: mcdp-ensembles-transition}). 
\parag{Ensembles:} When $\vS$ goes to infinity, \textit{Masks IoU}  goes to zero and Masksembles starts behaving like Ensembles that uses non-overlapping masks.  

\parag{Dropout:} As $\vN$ increases, each individual mask configuration becomes less and less likely to be picked during training. In the limit where $\vN$ goes to infinity, we reproduce the situation where each mask is seen only once, which is equivalent to MC-Dropout. In this case, the dropout rate would be $1 - \frac{1}{\vS}$.

\newcommand\toysize{0.16}
\renewcommand{\arraystretch}{0.0}
\begin{figure}[h]
\centering
\hspace*{-0.3cm}
\begin{tabular}{@{}c@{}c@{}c@{}}
  \includegraphics[width=\toysize\textwidth]{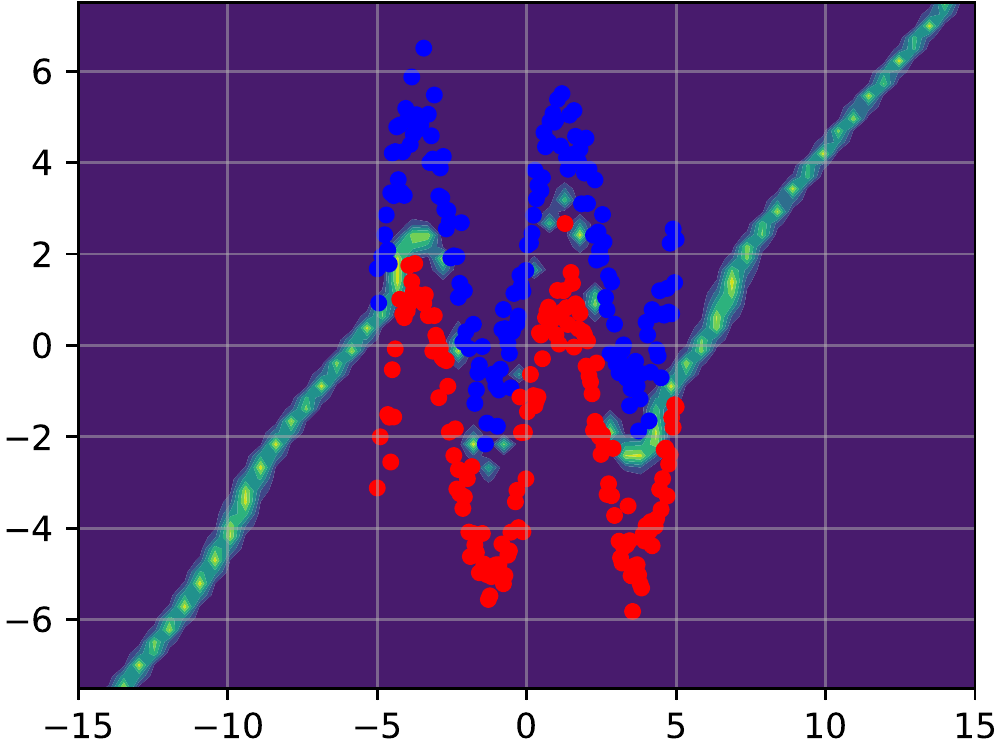} &   \includegraphics[width=\toysize\textwidth]{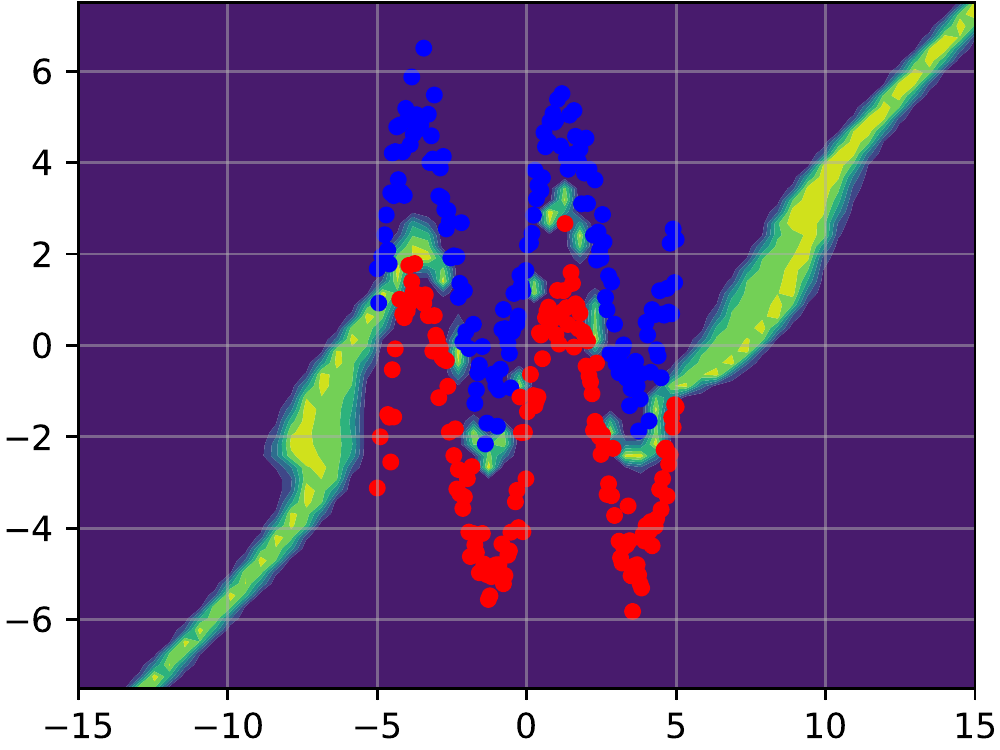}  & \includegraphics[width=\toysize\textwidth]{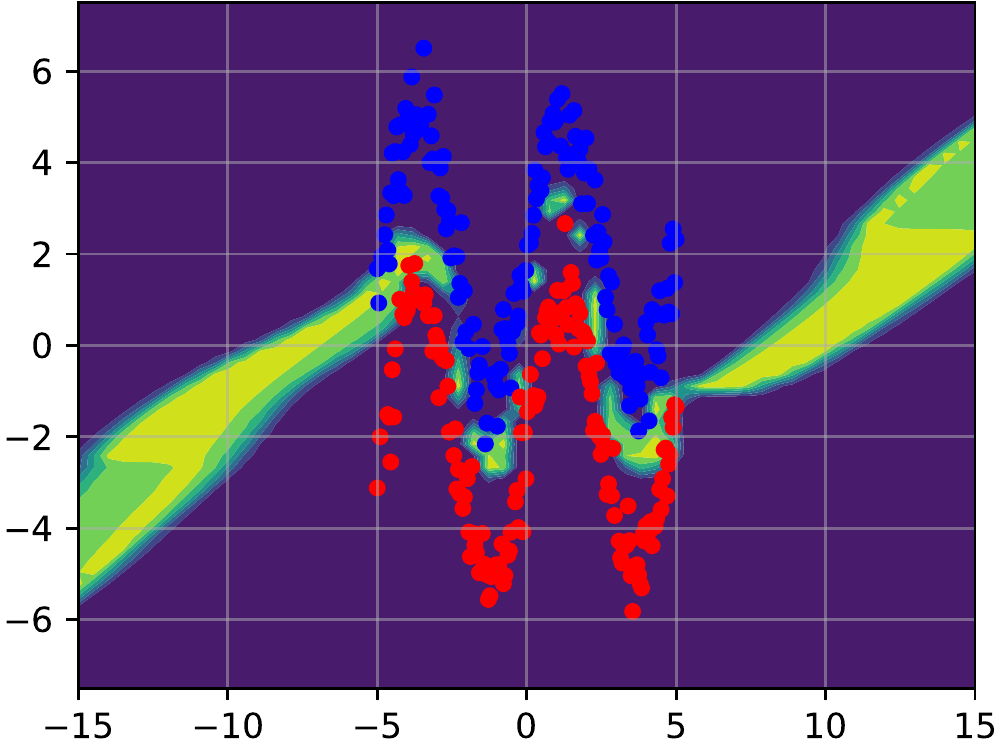} \\
  (a) & (b) & (c) \\[6pt]
   \includegraphics[width=\toysize\textwidth]{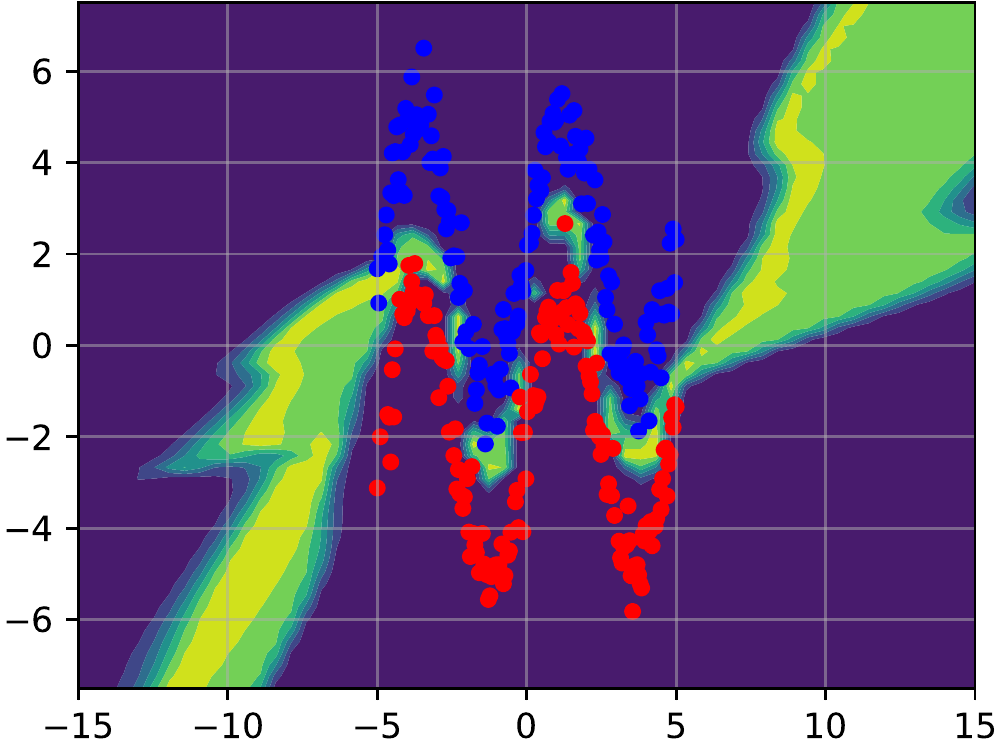} & \includegraphics[width=\toysize\textwidth]{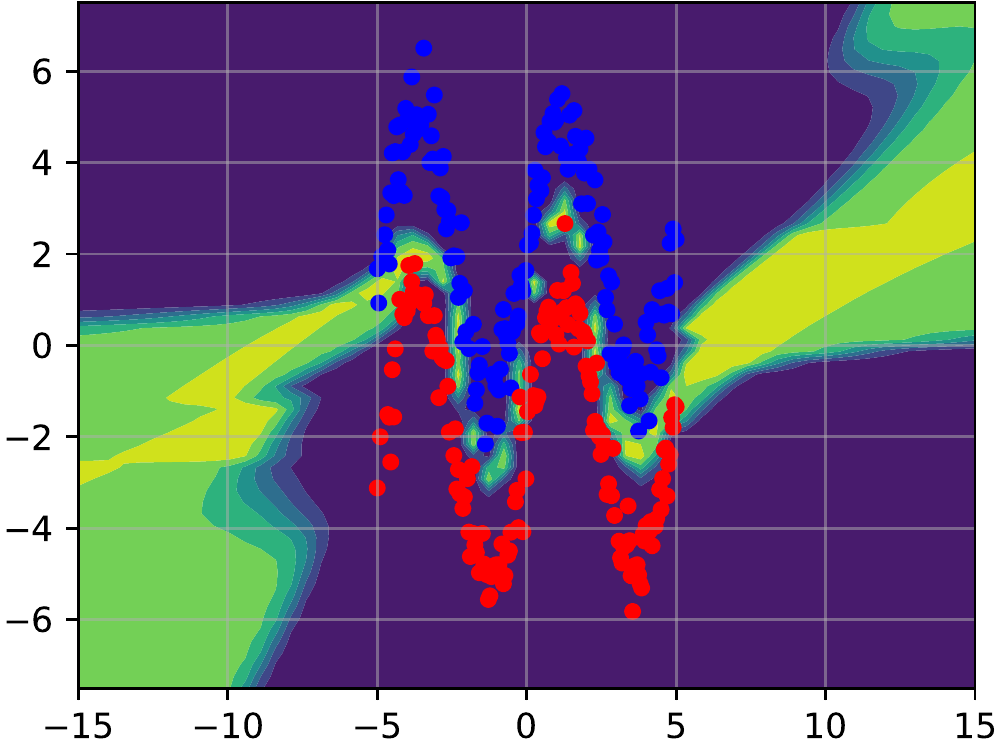} & \includegraphics[width=\toysize\textwidth]{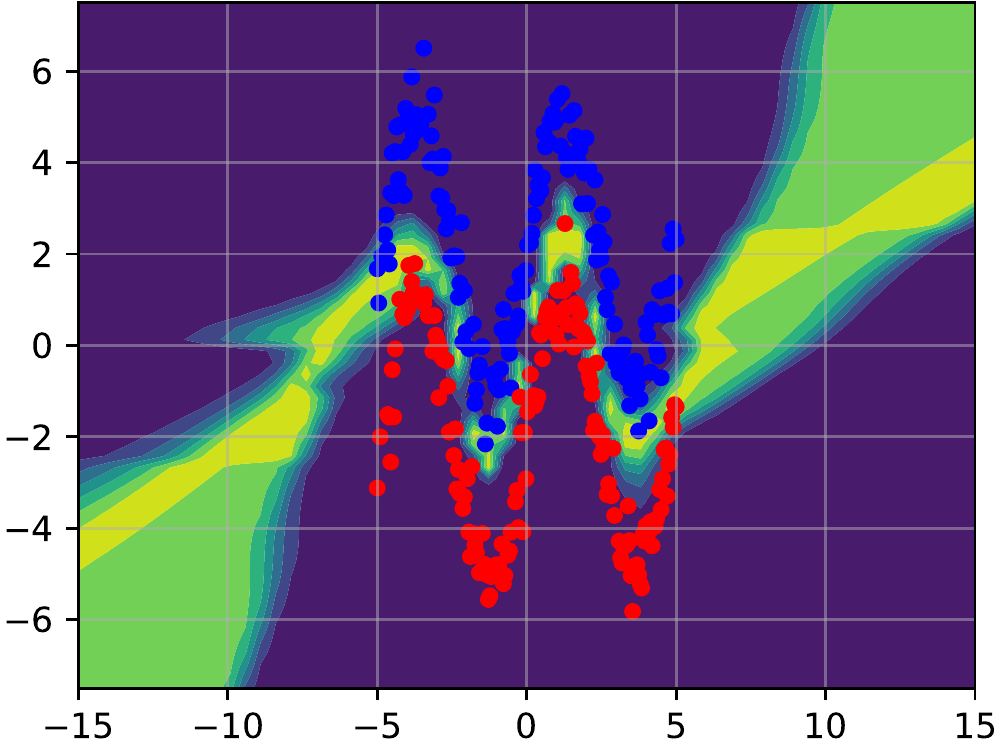} \\[3pt]
   (c) & (e) & (f) \\[6pt]
\end{tabular}
\caption{\small {\bf Single Model to Ensembles transition via Masksembles}. The task is to classify the red vs blue data points drawn in the range $x \in [-5, 5]$ from two sinusoidal functions with added Gaussian noise. The background color depicts the entropy assigned by different models to individual points on the grid, low in blue and high in yellow. \textbf{(a)} Single model, \textbf{(b) - (e)} Masksembles models with $\vN = 4$, $\vM = 100$, $\vS = [1.1, 2.0, 3.0, 10.0]$, \textbf{(f)} - Ensembles model.   For a fair comparison with ensembles, we used a fixed value for $\vM = 100$ when training Masksembles with different $\vS$, so that each Masksembles-model has the same 100 hidden-units capacity and the correlation between models decreases from (b) to (e). For high mask-overlap values, Masksembles behaves almost like a Single Model but starts behaving more and more like Ensembles as the overlap decreases.}
\label{fig: toy transition}
\end{figure}
\renewcommand{\arraystretch}{1.0}

\parag{Transition:}

In Fig.~\ref{fig: toy transition}, we use a simple classification problem to illustrate the span of behaviors Masksembles can cover. Specifically, we show that reducing mask overlap, that is, decreasing the correlation between binary masks, yields progressively more diverse predictions on out-of-training examples and generates more ensemble-like behavior.

\subsection{Implementation details}

%\paragraph{Computational Costs}
The only extra computation performed in Masksembles over a single 
network is cheap tensor-to-mask multiplication. Therefore, compared to convolutional and fully connected layers,  this product induces negligible computational overhead. 
Furthermore, we rely on similar optimizations as 
in~\cite{wen2020batchensemble} to make our implementation
more efficient.

\section{Experiments}
\label{sec: experiments}

In this section, we first introduce the evaluation metrics
which we use to quantitatively compare Masksembles against MC-Dropout and Ensembles. We then compare our method with the baselines on two broadly used benchmark datasets CIFAR10~\cite{krizhevsky2009learning} and ImageNet~\cite{deng2009imagenet}.

Note that we did not use any complex data augmentation 
schemes, such as AugMix~\cite{hendrycks2019augmix}, 
Mixup~\cite{zhang2017mixup} or adversarial 
training~\cite{goodfellow2014explaining}, in order to avoid 
entangling their effects with those of Masksembles. 
Ultimately, these techniques are complementary to ours and could easily be used together.

\subsection{Metrics}

Given a classifier with parameters $\vtheta$, let $p(y | \vx ; \vtheta)$ be the predicted probability for a sample represented 
by features $\vx$ to have a label $y$, and let $p_{gt}(y | \vx)$ 
denote the true conditional distribution over the labels
given the features. 
We use several different metrics to analyze how 
close $p(y | \vx ; 
\vtheta)$ is to $p_{gt}(y | \vx)$ and to compare our method to others, in terms of both accuracy and quality 
of uncertainty estimation. 

\parag{Accuracy.} For classification tasks, we use the standard classification accuracy, i.e. the percentage of correctly classified samples on the test set.

\parag{Entropy (ENT).} It is one of the most popular measures used to quantify uncertainty~\cite{lakshminarayanan2017simple,malinin2018predictive}. 
For classification tasks it is defined as
\vspace{-0.2cm}
\begin{equation*}
-\sum_{c=1}^{N_c} p(y = c | \vx ; \vtheta) \log{p( y = c | \vx ; \vtheta)} \;
\end{equation*}
where $N_c$ is the number of classes.

\parag{Expected Calibration Error (ECE).} ECE quantifies the quality of uncertainty estimates specifically for in-domain uncertainty. A model is considered well-calibrated if its predicted distribution over labels $p_(y | \vx ; \vtheta)$ is close to the true distribution $p_{gt}(y | \vx)$. 
As in~\cite{guo2017calibration}, we define ECE to be the average discrepancy between the predicted and real data distribution:
\vspace{-0.2cm}
\begin{equation*}
\mathbb{E}_{\vx,y\sim p_{gt}}\left|p(y | \vx ; \vtheta ) - p_{gt}(y | \vx)\right| \; ,
\end{equation*}
which is a common metric for calibration evaluation.

\parag{Out-of-Distribution Detection (OOD ROC / PR).} 
Domain shift occurs when the features in the training data do 
not follow the same probability distribution as those in the 
test distribution. 
Detecting out-of-distribution samples is an important task, and uncertainty estimation can be used for this purpose under the assumption that our model returns higher uncertainty estimates for such samples.
To this end, we use the uncertainty measure
produced by our models as a classification score that determines
whether a sample is in-domain or not. We then apply standard 
detection metrics, ROC and PR AUCs, to quantify the performance of 
our model.

\parag{Model Size} This corresponds to the \textit{Total model size} defined in Section~\ref{sec:transition}. 
A major motivation for Masksembles is to match the performance of Deep Ensembles while using a smaller model that requires significantly less memory. 
We use a total number of weights that parameterize our models as a 
proxy for that. In addition to the model size, we also report the 
total GPU time used to train any particular model.

\subsection{CIFAR10}
\label{sec:cifar}

CIFAR10 is among the most popular benchmarks for uncertainty estimation. 
The dataset is relatively small-scale, and very large models tend to overfit the data easily.
We, therefore, picked the Wide-ResNet-16-4~\cite{zagoruyko2016wide} for
our experiments and used training procedures similar to the
original paper. 
Following~\cite{Gal16}, we place dropout layers before all the 
convolutional and fully connected layers in the MC-Dropout model. 
Masksembles layers are added exactly the same way.

\parag{Accuracy and ECE.} 
One of the standard ways to assess the robustness
of a model is to evaluate the dependency of its 
performance with respect to inputs perturbations~\cite{hendrycks2019benchmarking, ovadia2019can}.
In our experiments, we investigate the behaviour
of all the considered models under the influence of domain shifts induced by typical sources of noise. 

Following~\cite{wen2020batchensemble,ovadia2019can}, 
we evaluate model accuracy and ECE on a corrupted version of CIFAR10~\cite{hendrycks2019benchmarking}. 
Namely, we consider 19 different ways to artificially
corrupted the images and 5 different levels of severity
for each of those corruptions.

\begin{figure}
    \centering
    \includegraphics[width=0.48\textwidth]{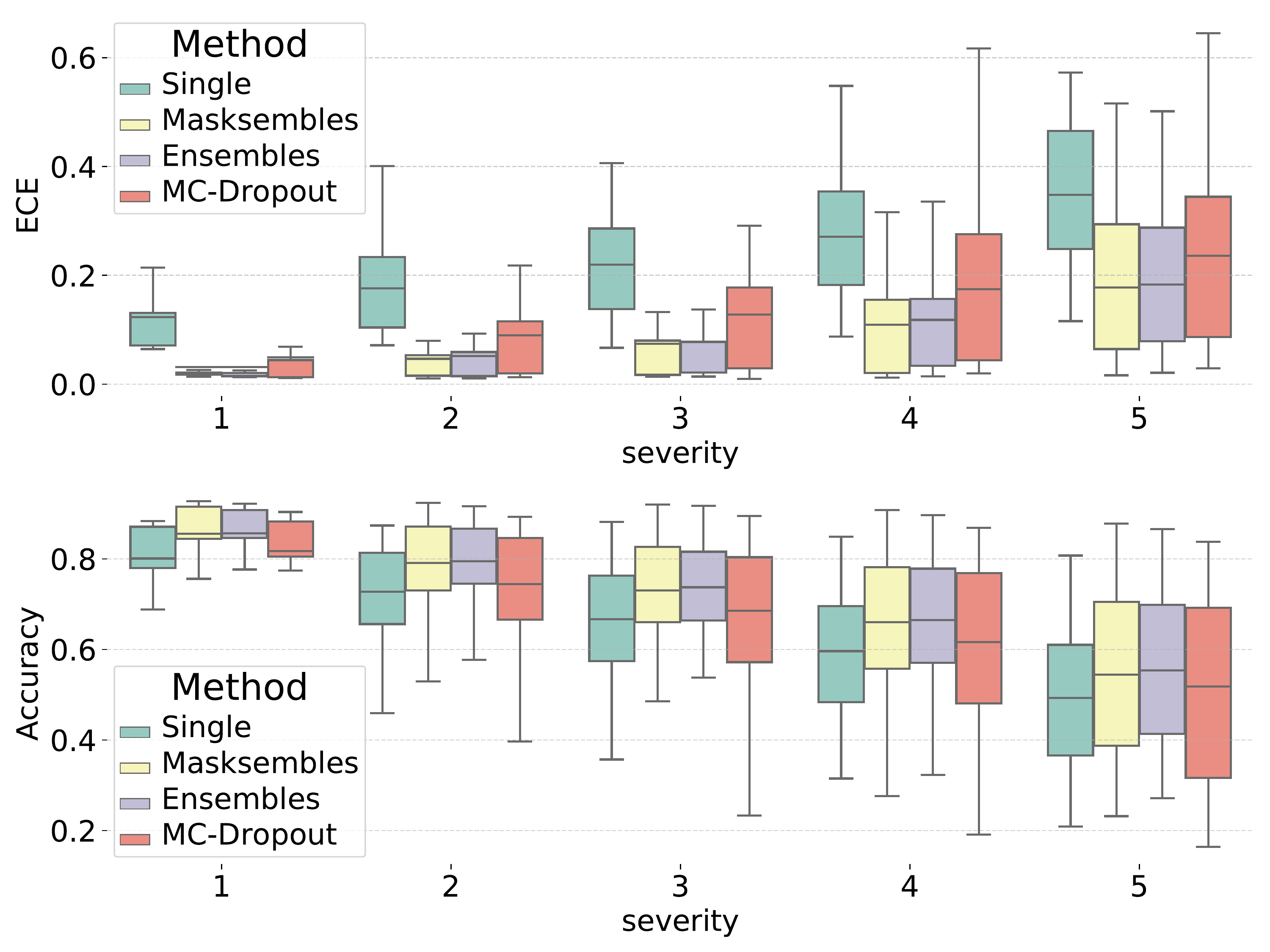}
    \caption{\small {\bf CIFAR-10 results on corrupted images.} \textcolor[HTML]{D6BA73}{\textbf{Masksembles}} and \textcolor[HTML]{ea8e83}{\textbf{MC-Dropout}} models with tuned (on validation set) dropout rate parameter values, \textcolor[HTML]{d1bed6}{\textbf{Ensemble}} of independently trained models and \textcolor[HTML]{96cac1}{\textbf{Single}} model. All approaches, except Single, use 4 models whose predictions are averaged.}
    \label{fig: cifar10 corrupted}
\end{figure}

We train all the models on the original uncorrupted CIFAR10 
training set and test them both on the original images 
and their corrupted versions.
In Table~\ref{tab:cifar10 results}, we report our results on the original uncorrupted images and in Fig.~\ref{fig: cifar10 corrupted} the results on the noisy ones. 
Each box represents the first and third quantiles of accuracy and ECE while the error bars denote the minimum and maximum values. 
%
% TODO: if we have time, we should have a separate section
% for baselines
We tested four different approaches, that is, a single network,
MC-Dropout, Ensembles, and our Masksembles.  
Unsurprisingly, as the severity of the perturbations increases, 
using a single network is the least robust approach and degrades 
fastest. Masksembles performs on par with Ensembles and 
consistently outperforms MC-dropout, even though we extensively 
tuned MC-Dropout for the best possible performance by
doing a grid search on the dropout rate (which turned out to be $p \approx 0.1$).

% In these experiments, we trained Masksembles using a set of 4 masks such that each submodel has similar capacity as a single model and shares on average 10\% of its weights with any other. The resulting model size is about $2.3$ times that of the single model while Ensembles is $4$ times that. 

\begin{figure}
    \centering
    \hspace*{-0.2cm}
    \includegraphics[width=0.5\textwidth]{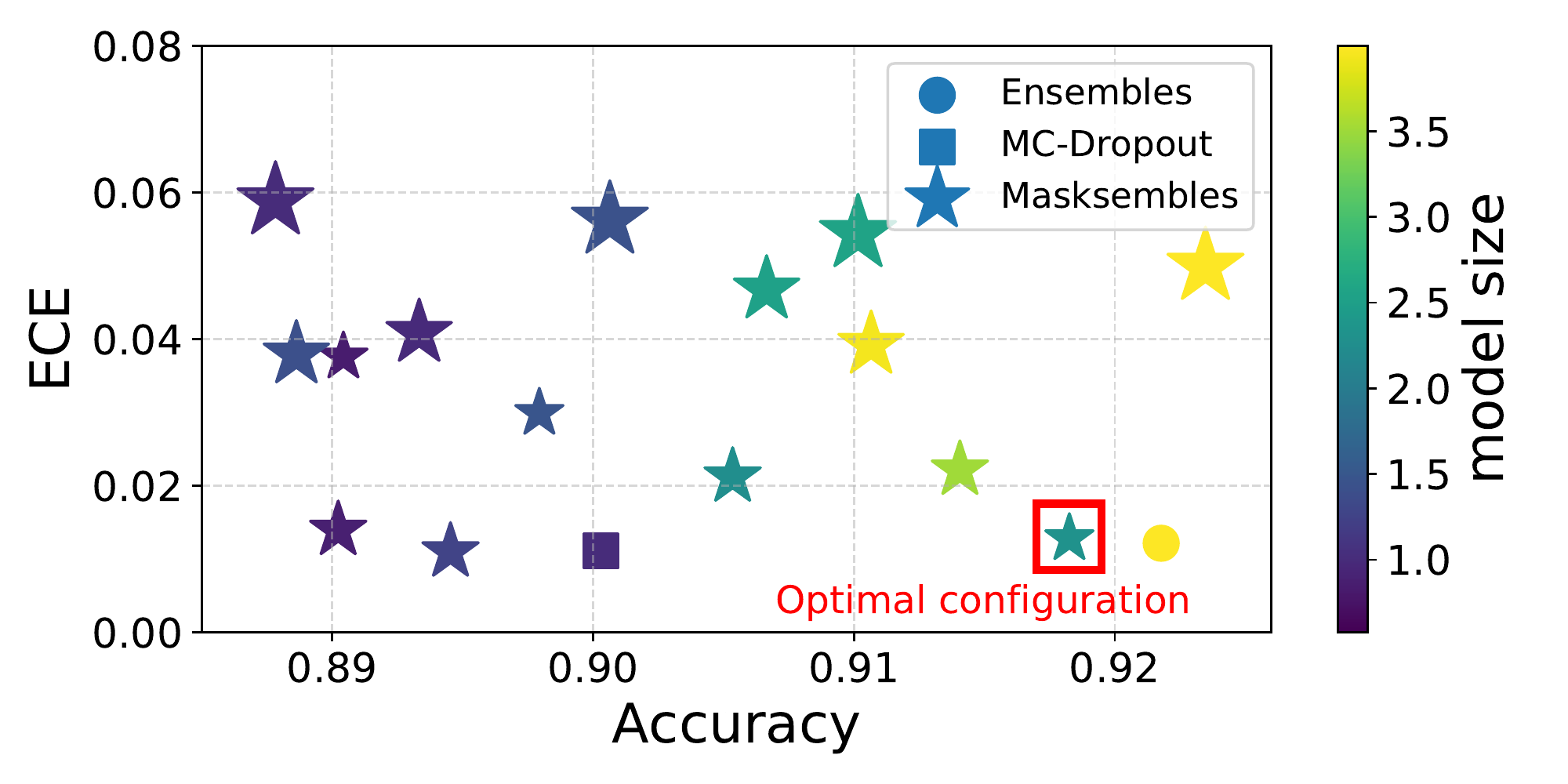}
    % \vspace{-3mm}
    \caption{\small {\bf Spanning the space of behaviors}. Models in the bottom right corner are better. The color represents the relative size of the model compared to a single model. The size of Masksembles markers denotes mask overlap between $1.0$ and $0.2$.}
    \label{fig: acc vs ece}
\end{figure}

In these experiments, Masksembles was trained with a fixed number of $\vN = 4$ masks. We varied both $\vS$ in the range [$1.0$, $5.0$] and $\vM$ in the range [$0.3C$, $2.0C$],  where $C$ is the number of channels in the original model. Fig.~\ref{fig: acc vs ece} depicts the resulting range of behaviors. 
The 2D coordinates of the markers depict their accuracy and ECE, 
their colors - the corresponding model size, and the size of the star markers denotes the value of the mask-overlap. For comparison purposes, we also display MC-dropout and Ensembles results in a similar manner, simply replacing the star with a square and a circle, respectively. As can be seen, the optimal Masksemble configuration depicted by the green star in the lower right corner delivers performances similar to those of Ensembles for a model size that is almost half the size. This is the one we also used in the experiments depicted by Fig.~\ref{fig: cifar10 corrupted}.

\begin{figure}
    \centering
    \hspace*{-0.2cm}
    \includegraphics[width=0.48\textwidth]{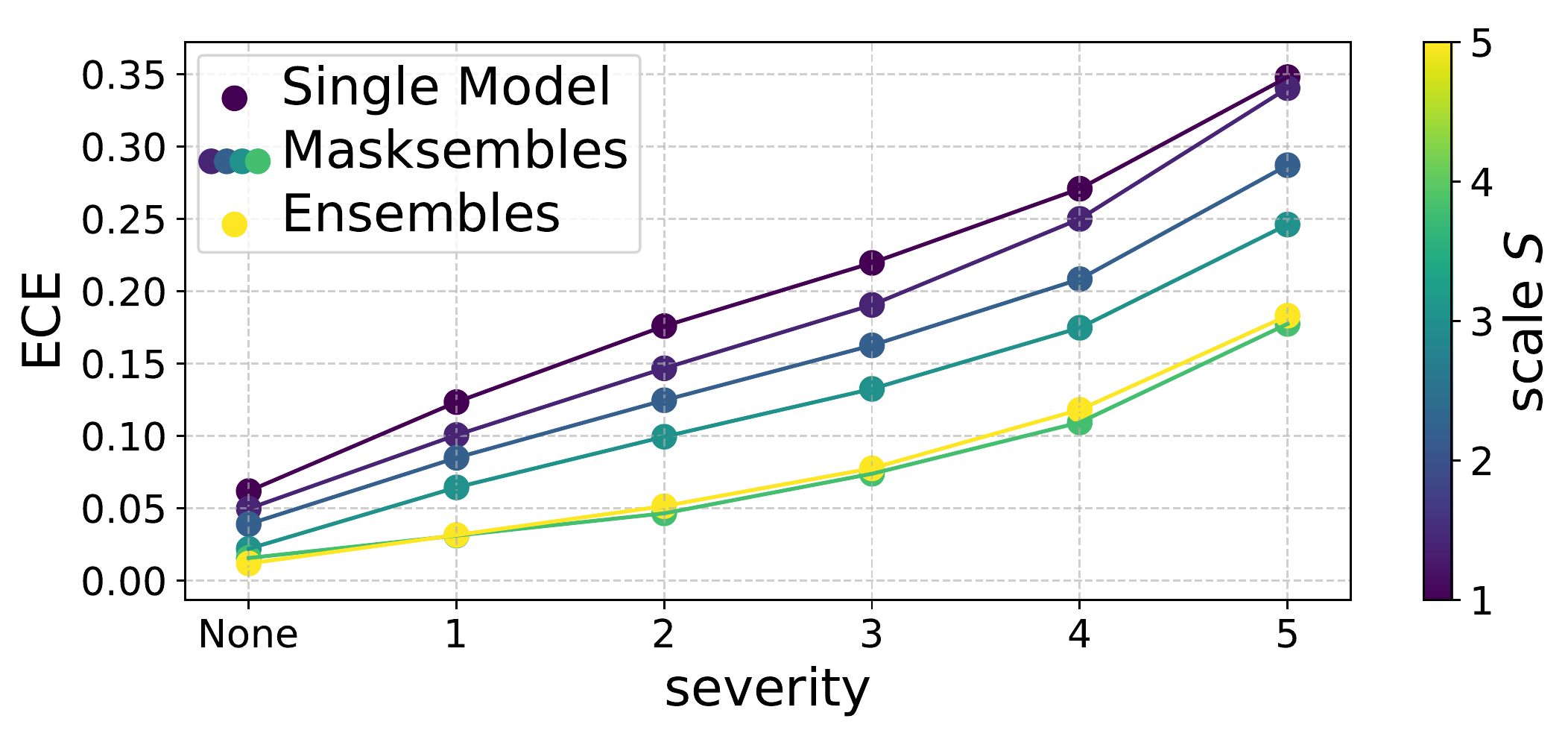}
    % \vspace{-5mm}
    \caption{\small {\bf CIFAR10 ECE}. ECE as a function of severity of image corruptions. Each Masksembles curve corresponds to a different $\vS$. In this experiment we force $\vM \times \vS$ to remain constant and equal to the original Wide-ResNet channels number and we vary $\vS$ in the range [$1., 1.1, 1.4, 2.0, 3.0$]. Larger values of $\vS$ correspond to more ensembles-like behavior.}
    \label{fig: cifar ece fixed}
\end{figure}

In Fig.~\ref{fig: cifar ece fixed}, we chose the Masksembles parameters so that the model size always remains the same: We fix the number of masks $\vN = 4$ and vary $\vS$ and $\vM$ so that the \textit{total model size} remains constant, which means that a lower mask overlap results in a lower \textit{capacity} for each individual model. This procedure is important in practice because it enables us to take a large network, such as ResNet, and add Masksembles layers as we would add Dropout layer {\it without} having to change the rest of the architecture.

For this set of experiments we obtain a consistent improvement in calibration quality by reducing mask overlap and achieve ensembles-like performance with $\vS = 5$.

\renewcommand{\arraystretch}{0.0}
\begin{figure}[h]
\centering
% \vspace*{-0.4cm}
\begin{tabular}{@{}c@{}c@{}c@{}}
  \includegraphics[width=0.16\textwidth]{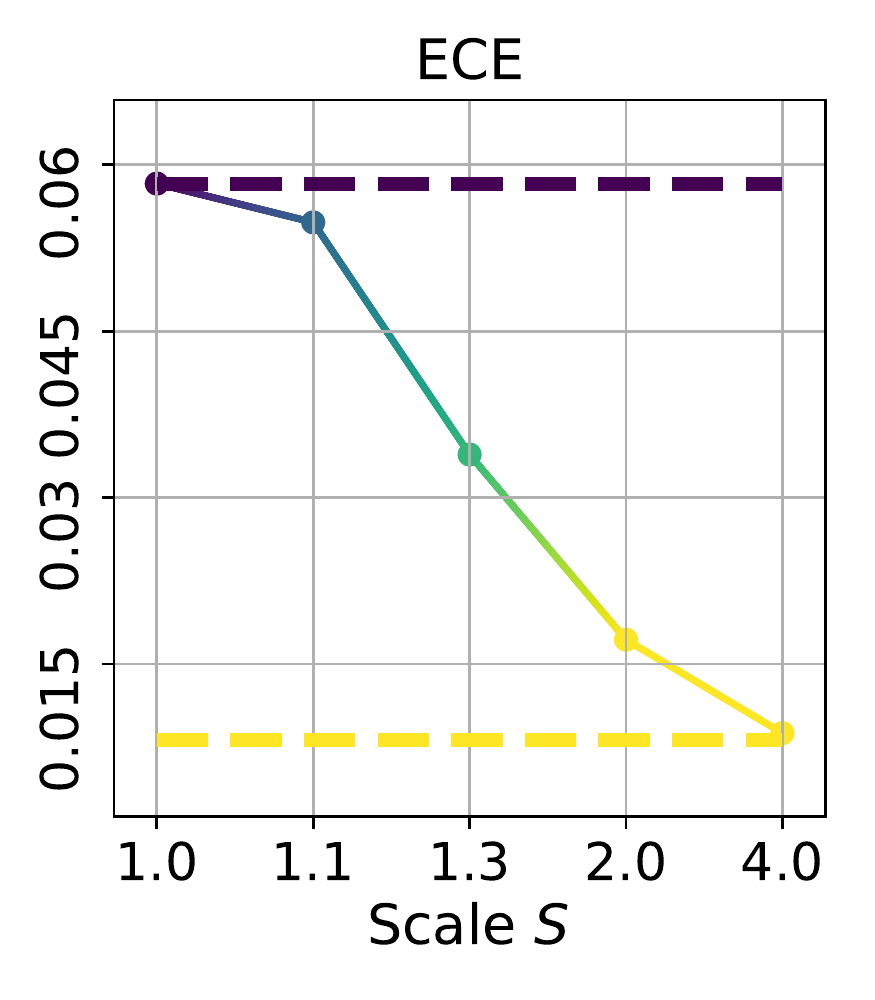} &
  \includegraphics[width=0.16\textwidth]{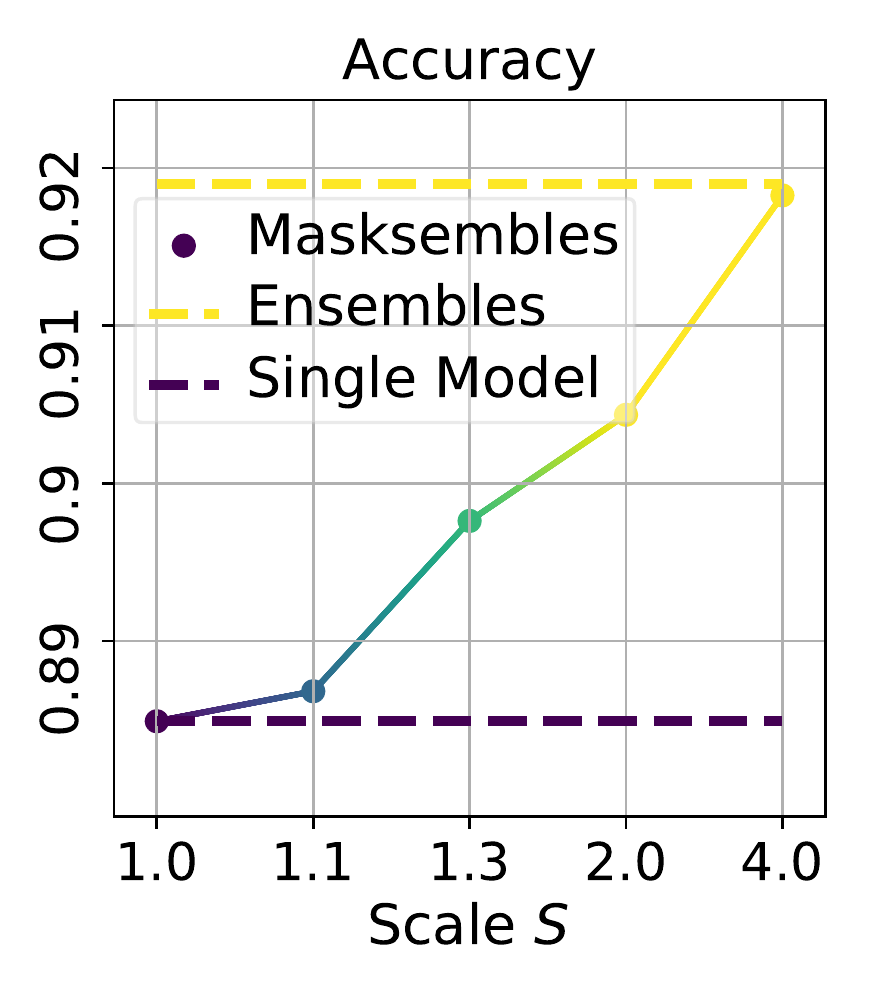} &
  \includegraphics[width=0.16\textwidth]{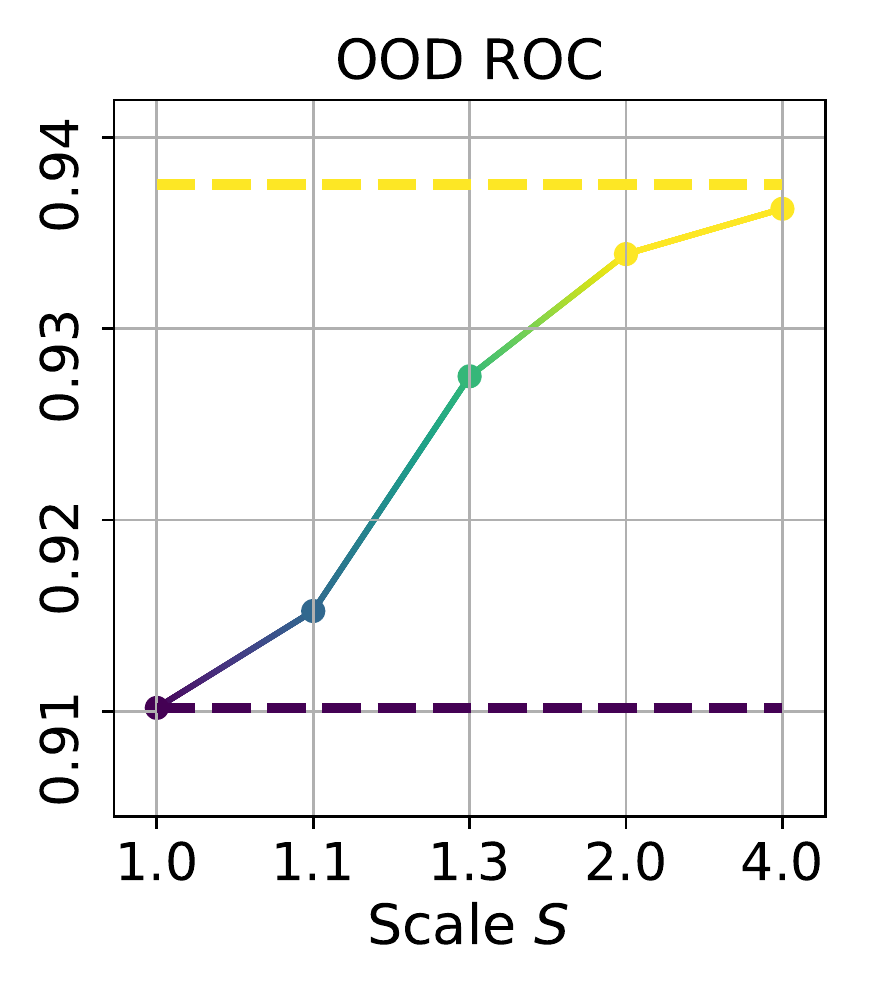} \\
\end{tabular}
\caption{\small{\textbf{Fixed capacity for CIFAR10}. ECE, Accuracy and OOD as functions of the scale parameter $\vS$ with fixed $\vM$ and $\vN$. The colors represent different masks overlap values.}}
\label{fig: fixed M}
\end{figure}
\renewcommand{\arraystretch}{1.0}
% \vspace*{-0.3cm}

\parag{OOD ROC and PR.}

For OOD task, we followed the evaluation protocol of~\cite{ciosek2019conservative}. 
Namely, we trained our models on CIFAR10, and then used CIFAR10 
test images as our in-distribution samples and images from the 
SVHN dataset~\cite{netzer2011reading} (which belong to classes that are not in CIFAR10), as our out-of-distribution samples. 
We report our OOD results in the two bottom rows of Table~\ref{tab:cifar10 results}. 
Again, performance of our method is similar to that of Ensembles at a fraction of the computational cost, 
and significantly better than that of a single network and MC-dropout.

\begin{table}

  \begin{center}

    \begin{tabular}{c|cccc} 
      & Single & MC-D & MaskE & NaiveE\\
      \hline
      Accuracy & 0.89 & 0.90 & \textbf{0.92} & \textbf{0.92}\\
      ECE & 0.06 & \textbf{0.01} & \textbf{0.01} & \textbf{0.01}\\
      \hline
      Size & 1x & 1x & 2.3x & 4x \\
      Time & 10m & 12m & 16m & 40m \\
      \hline
      OOD ROC & 0.91 & 0.92 & \textbf{0.94} & \textbf{0.94} \\
      OOD PR & 0.94 & 0.95 & \textbf{0.96} & \textbf{0.96} \\
    \end{tabular}
  \end{center}
  \caption{\small {\bf CIFAR10 results.} Each model uses 4 samples during inference and we used the uncorrupted versions of the images.}  
    \label{tab:cifar10 results}  
    \vspace{-0.3cm}
\end{table}

\parag{Fixed Capacity Model.}

In order to provide clear evidence for Single Model $\xleftrightarrow{}$ Ensembles transition of Masksembles we perform fixed capacity experiments and report the results in Fig.~\ref{fig: fixed M}.
Similarly to our previous experiments we set $\vN = 4$ and vary the scale parameter in range $\vS \in [1.0, 4.0]$, while keeping parameter $\vM$ fixed. 
These results confirm that Masksembles is able to span the entire spectrum of behaviours 
between different models: the key metrics gradually improve as $\vS$ is increased,
and the Masksembles's behavior (solid line) clearly transitions from 
Single Model to Ensembles behavior (dashed lines).

\subsection{ImageNet}

The ImageNet~\cite{deng2009imagenet} is another dataset
that is widely used to assess the ability of neural networks to 
produce uncertainty estimates.
For this dataset, we rely on the well-known ResNet-50~\cite{he2016deep} architecture as a base model, 
using the setup similar to that in~\cite{ashukha2020pitfalls}. 
Similarly to our CIFAR10 experiments, for the MC-Dropout and 
Masksembles models, we introduce dropout (masking) layers before 
every convolutional layer starting from \texttt{stage3} layers.

\begin{figure}
    \centering
    \hspace*{-0.2cm}
    \includegraphics[width=0.48\textwidth]{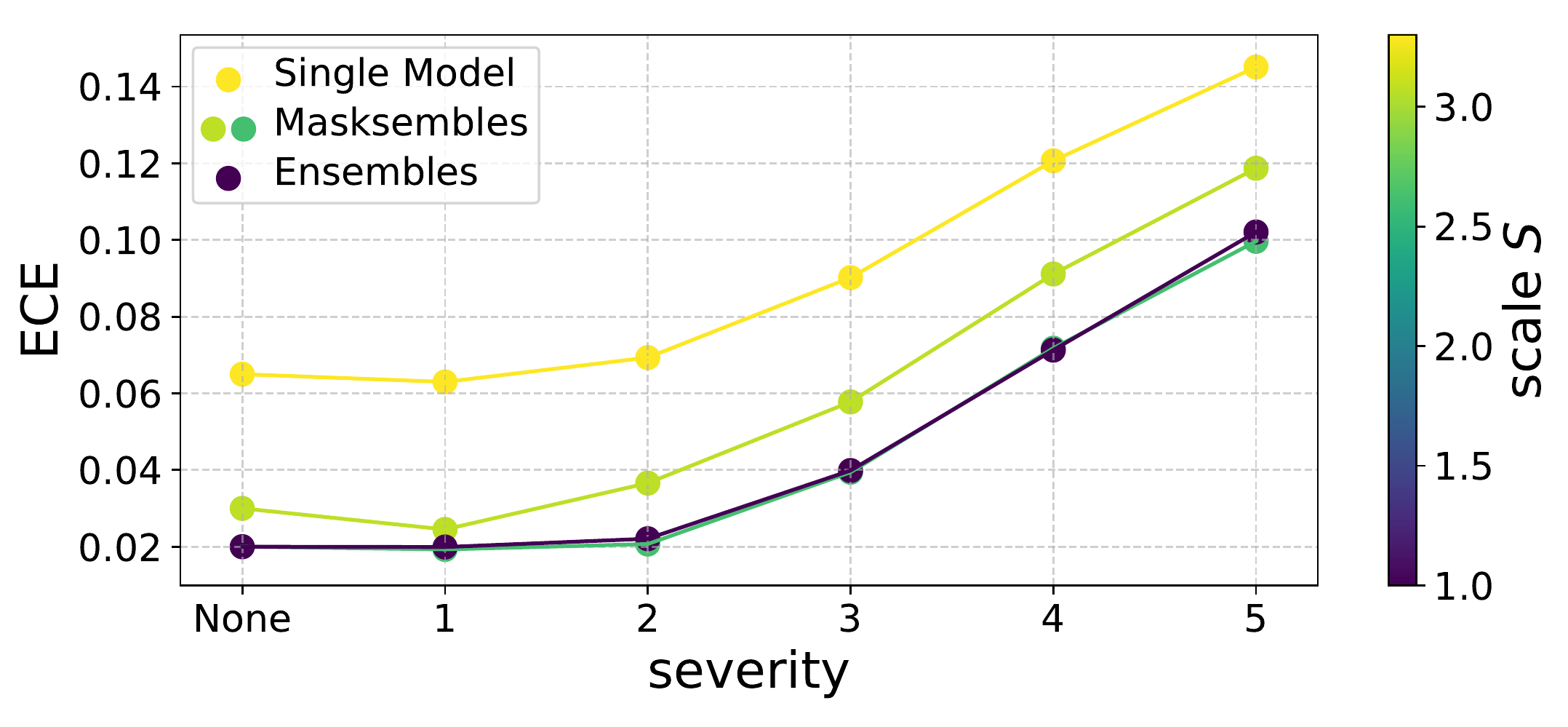}
    %\vspace{-5mm}
    \caption{\small {\bf ImageNet ECE}. ECE as a function  of severity of image corruptions. Each Masksembles curve corresponds to a different $\vS$. In this experiment we force $\vM \times \vS$ to be constant and equal to original ResNet-50 channels number.}
    \label{fig: imagenet ece fixed}
\end{figure}

\begin{figure}
    \centering
    \includegraphics[width=0.49\textwidth]{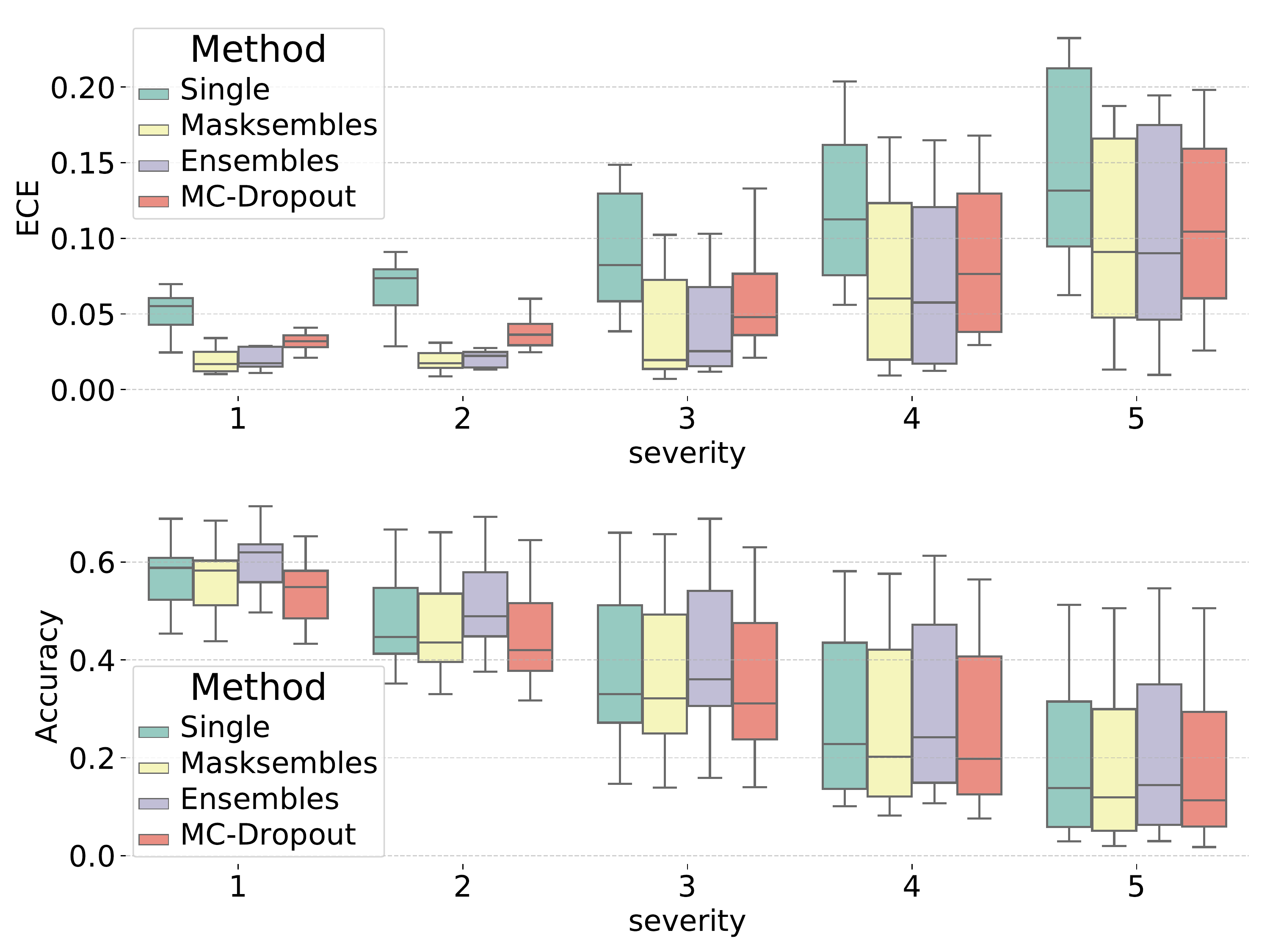}
    \caption{\small {\bf ImageNet results on corrupted images.}  \textcolor[HTML]{D6BA73}{\textbf{Masksembles}} and \textcolor[HTML]{ea8e83}{\textbf{MC-Dropout}} with tuned dropout rate parameter values, \textcolor[HTML]{d1bed6}{\textbf{Ensemble}} of independently trained models and \textcolor[HTML]{96cac1}{\textbf{Single}} model. All approaches, except \textbf{Single}, include 4 models, their predictions are averaged.}
    \label{fig:imageNet}
    \vspace{-0.2cm}
\end{figure}

We followed exactly the same evaluation protocol as in 
Section~\ref{sec:cifar} and report our results on the original 
images in Table~\ref{tab:imagenet reduced overlap} and on the corrupted ones in Fig.~\ref{fig:imageNet}. 
In Table~\ref{tab:imagenet reduced overlap} and Fig.~\ref{fig: imagenet ece fixed} we use the same procedure as for CIFAR10 to select Masksembles parameters so that the \textit{total model size} remains fixed. 

For this dataset, Masksembles demonstrates its best performance
for a choice of parameters corresponding to a \textit{Masks IoU} of 0.5. 
Similarly to our results on CIFAR10, the performance of Masksembles
both in terms of accuracy and ECE is very close to that of Ensembles and significantly better than that of MC-Dropout.
Note that our method achieves these results with 
a training time and memory consumption that is 4 (four) times 
smaller than that of Ensembles, and nearly the same as that of a 
single network. In terms of calibration quality on corrupted images, we achieve Ensemble-like performance starting from $\vS = 2$ and up.

\begin{table}
  \begin{center}
    \begin{tabular}{@{}p{6mm}|c|ccc|c|c} 
      & Single & \multicolumn{3}{c|}{Masksembles} & NaiveE & MC-DP\\
      \hline
      %\cellcolor[gray]{0.8}
      Acc. & 0.71 & 0.72 &  0.71 & 0.70 & 0.74 & 0.69 \\
      ECE & 0.07 & 0.03 & 0.02 & 0.02 & 0.02 & 0.03\\
      \hline
      IoU & 1.0 & 0.7 & 0.3 & 0.2 & - & -\\
      Time & 50h & 55h & 60h & 70h & 200h & 80h\\
    \end{tabular}
  \end{center}
  \caption{\small {\bf ImageNet results.}  Accuracy and ECE results for \textbf{Single}, \textbf{Masksembles} models using masks overlapping values (0.7, 0.3, 0.2), \textbf{Ensembles}, and \textbf{MC-Dropout}. All the models have the same size as the \textbf{Single model}, except for \textbf{Ensembles} which is 4x times larger.}
    \label{tab:imagenet reduced overlap}  
\end{table}

\section{Conclusion}
\vspace{-0.13cm}

In this work, we introduced Masksembles, a new approach to uncertainty estimation in deep neural networks. 
Instead of using a fixed number of independently trained models as in Ensembles or drawing a new set of random binary masks at each training step as in MC-Dropout, 
Masksembles predefines a pool of binary masks with a controllable overlap and
stochastically iterates through them. 
By changing the parameters that control the mask generation
process, we can span a range of behaviors between those of 
MC-Dropout and Ensembles. 
This allows us to identify model configurations that 
provide a useful trade-off between the high-quality 
uncertainty estimates of Deep Ensembles at a high 
computational cost and the lower performance
of MC-Dropout at a lower computational cost.
In fact, our experiments demonstrate that we can achieve the
performance on par with that of Deep Ensembles at a fraction
of the cost. 
Because Masksembles is easy to implement, we believe it can serve as a drop-in replacement for MC-Dropout. 
In future work, we will explore ways to accelerate
Masksembles by exploiting the structure of the masks,
and apply our method to tasks 
that require a scalable uncertainty estimation
method, in particular reinforcement learning and Bayesian 
optimization.

%
%We showed that dropout and Ensembles can be seen as extreme cases of Masksemble models. Therefore, this new approach provides a new way to optimize the performance - efficiency trade-off between those approaches. We saw that, with a good choice of MaskSembles parameters, it is often possible to obtain similar performances to Ensembles, at a fraction of the computational cost, or to obtain radically better performances than dropout, at the same cost.
%

%
%Furthermore, by creating and studying continuum between drop-out and ensembles, we believe that our work can contribute to a better theoretical understanding of these methods.

\clearpage

{\small
\bibliographystyle{ieee_fullname}
\bibliography{bib/string,bib/vision,bib/learning,bib/new}
}

\clearpage
\appendix
\section{Supplementary Material}

In this appendix, we study in more detail the relationship between the choice for  $\vN$, $\vM$, and $\vS$ parameters of Section~3.2 that control the mask generation process and the resulting model size, mask correlation, and models' diversity. 

\subsection{Expected models size}

We first provide a formal derivation for the expected size of a generated model given the $\vN, \vM, \vS$ parameters. Let us consider $\vN$ vectors of size $\vM \times \vS$ filled with zeros, and then in each of those vectors, we randomly set $\vM$ of these elements to be 1. Let $\zeta_{ij}$ be a random variable that represents whether or not $j$-th position in $i$-th vector is equal to one.
\begin{equation}
    \mathbb{P}(\zeta_{ij} = 1) = \frac{\vM}{\vM \times \vS} = \frac{1}{\vS} \; .
\end{equation}
This arises from the fact that there are $\binom{\vM \times \vS}{\vM}$ ways to chose $\vM$ positions from $\vM \times \vS$ places and only $\binom{\vM \times \vS - 1}{\vM - 1}$ such configurations where $j$-th position is fixed to be one, therefore 
\begin{equation}
    \begin{split}
        \mathbb{P}(\zeta_{ij} = 1) &= \binom{\vM \times \vS - 1}{\vM - 1} \cdot \binom{\vM \times \vS}{\vM}^{-1} \\
        & = \frac{\vM}{\vM \times \vS} = \frac{1}{\vS}\;
    \end{split}
\end{equation}

Let $\epsilon_j = [\sum_{i=1}^{\vN} \zeta_{ij} > 0]$ be the random variable representing whether at least one 1 appears in the $j$-th position among all generated vectors. Given that $\mathbb{P}(\zeta_{ij} = 0) = 1 - \frac{1}{\vS}$ and $\mathbb{P}(\overline{X}) = 1-\mathbb{P}(X)$, then probability of $\epsilon_j$ can be written as
\begin{equation}
        \mathbb{P}(\epsilon_j = 1) = 1 - (1 - 1/\vS)^{\vN} \; .
\end{equation}

To compute the expected model size, we compute the expectation of $\epsilon_j$ sum, since effectively it represents expected number of features that one will acquire after a trimming procedure: 

\begin{equation}
    \begin{split}
        Size(\vN, \vM, \vS) &= \mathbb{E}\sum_{i=1}^{\vM \times \vS} \epsilon_j = \sum_{i=1}^{\vM \times \vS}\mathbb{E}\epsilon_j \\
        & = \vM \times \vS \left[1 - (1 - 1/\vS)^{\vN}\right]
    \end{split}
\end{equation}

\subsection{Average IoU}

\begin{figure}[h]
    \centering
    \includegraphics[width=0.45\textwidth]{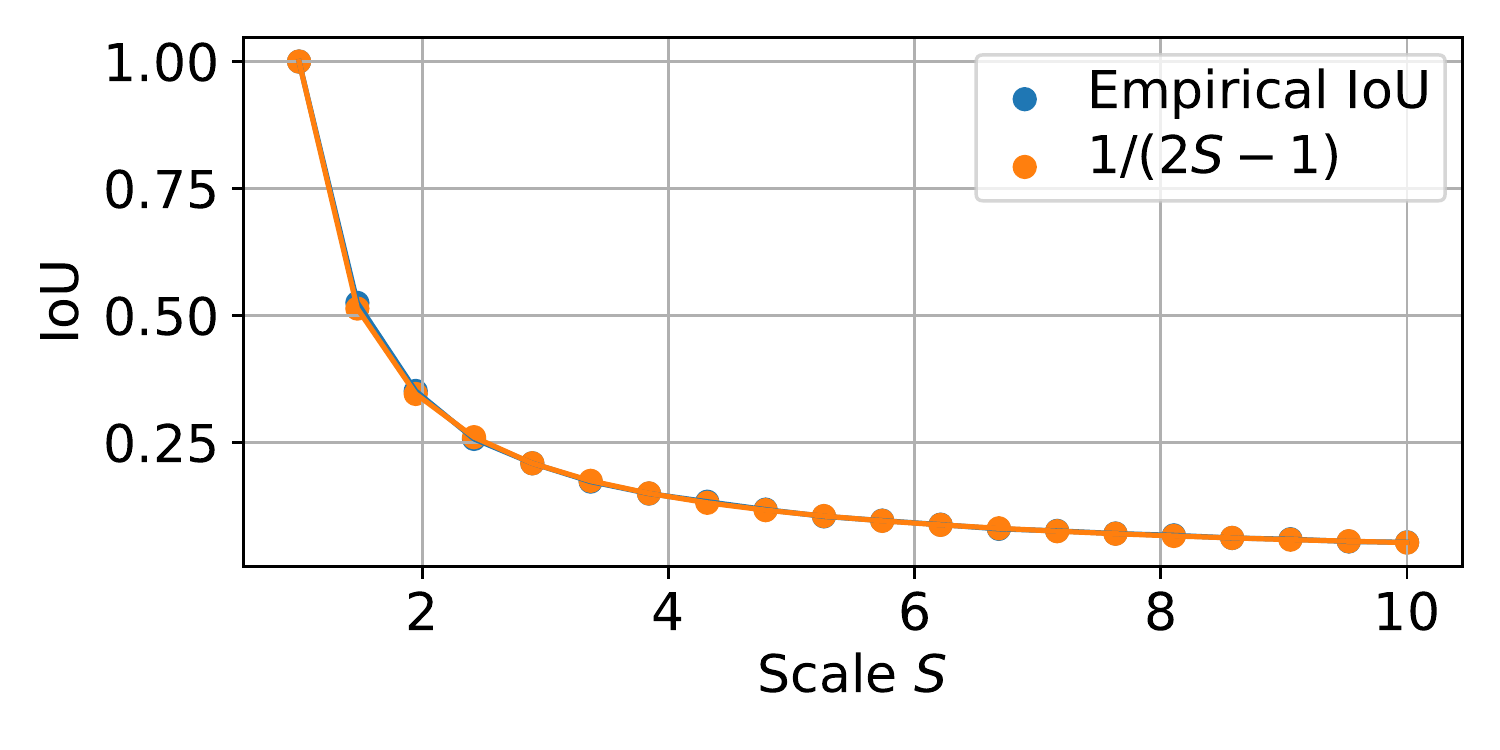}
    \caption{\small {\bf Empirical and Analytical IoU}. The plot represents how close real IoU of generated masks are and analytical approximation for it. In wide range of $\vS$ values we demonstrate a close match of considered quantities.}
    \label{fig:meanIoU}
\end{figure}

We now justify our $\frac{1}{2\vS - 1}$ approximation for the average mask IoU. Let us consider two vectors produced by our mask generation algorithm. As above, let $\zeta_{1j}$ and $\zeta_{2j}$ be random variables that represent the value in the $j$-th position in the first and second masks, respectively. Starting from the standard definition of the IoU, we estimate the average intersection $I$ of two such masks as the number of common ones:
\begin{align}
        \mathbb{E}{I} & = \mathbb{E}\sum_{j=1}^{\vM \times \vS} \zeta_{1j} \cdot \zeta_{2j} = \sum_{j=1}^{\vM \times \vS} \mathbb{E}\zeta_{1j} \cdot \mathbb{E}\zeta_{2j} \; \nonumber\\
        & = \vM \times \vS \cdot \frac{1}{\vS^2} = \frac{\vM}{\vS} \; .
\end{align}
Given two masks with $\vM$ ones each and intersection $I$, their IoU is $\frac{I}{2\vM - I}$. Therefore, a simple approximation for expected IoU is
\begin{equation}
    \begin{split}
        \mathbb{E}\left[\text{IoU}\right] & = \mathbb{E}\left[\frac{I}{2\vM - I}\right] \approx \frac{\mathbb{E}I}{2\vM - \mathbb{E}I} \\
        & = \frac{\vM / \vS}{2\vM - \vM / \vS} = \frac{1}{2\vS - 1}
    \end{split}
\end{equation}
In Fig.~\ref{fig:meanIoU}, we plot this value as a function of $S$ along with empirical values and the agreement is excellent. 

\subsection{Diversity analysis}

\begin{figure}[h]
    \centering
    \includegraphics[width=0.45\textwidth]{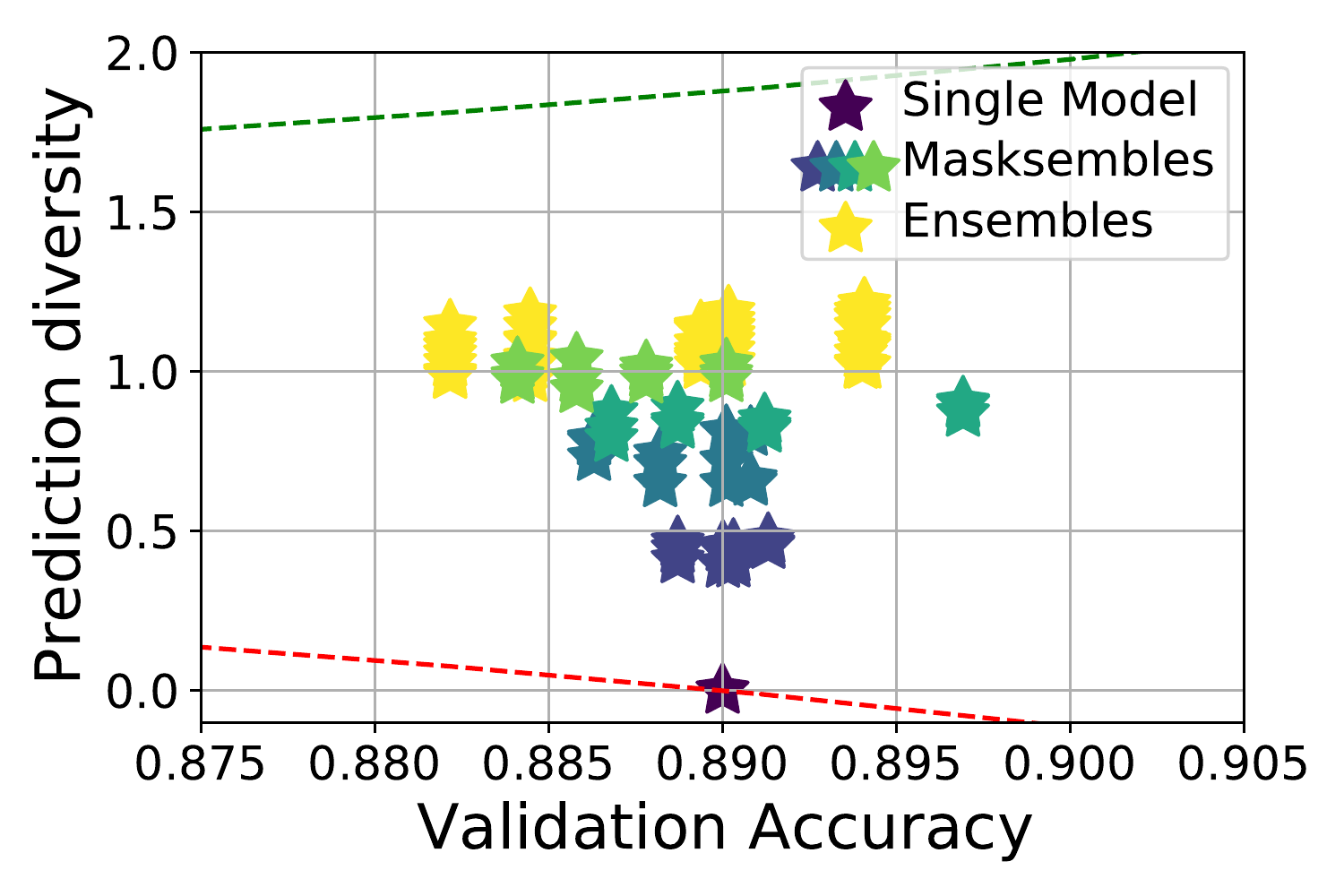}
    \caption{\small {\bf Diversity vs Accuracy trade-off for Cifar10}. Every point is associated with a pair of compared models. Larger diversity values correspond to less correlated models. \textcolor[HTML]{008000}{\textbf{Green}} and \textcolor[HTML]{ff0000}{\textbf{Red}} dashed lines represent the worst and the best theoretically possible diversity for a fixed accuracy.}
    \label{fig: diversity vs ece}
\end{figure}

It is known that less correlated ensembles of models deliver better performance, produce more accurate predictions~\cite{lakshminarayanan2017simple, perrone1992networks, hansen1990NNensembles}, and demonstrate lower calibration error~\cite{ovadia2019can}. Hence, a strength of Masksembles is that it provides the ability to control how correlated models within Masksembles are by adjusting the $\vN$, $\vM$, and $\vS$ parameters.

We now perform a diversity analysis of Masksembles models using the metric of~\cite{fort2019deep}. It involves comparing two different models trained on the same data in terms of how different their predictions are. Measuring fraction of the test data points on which predictions of models disagree, the \textit{diversity}, and normalizing it by the models error rate, we write
\begin{equation}
    \begin{split}
        \text{Diversity} = \frac{\text{fraction of disagreed labels}}{1.0 - \text{accuracy}} \; 
    \end{split}
\end{equation}
Plotting this diversity against accuracy allows us to look at our models from a bias-variance trade-off perspective. Since we want our models to be less correlated---larger \textit{diversity}--- and at the same time to be accurate, the upper-right corner of Fig.~\ref{fig: diversity vs ece} is where the best models should be.

For this experiment we used a single trained model; several Masksembles models with $\vN = 4$, fixed $\vM$ to have the same capacity as the  single one and varying $\vS $ in [$2, 3, 4, 5$]; ensembles model with  $4$ members. Fig.~\ref{fig: diversity vs ece} depicts the results on CIFAR10 dataset. Ensembles have the largest diversity  but Masksembles gives us the ability to achieve very similar results by controlling its parameters. The single model has $0$ diversity by the definition.

\subsection{Calibration Plots}

Since \textit{Expected Calibration Error} (ECE) provides only limited and aggregated information about model's calibration therefore in this section we present full calibration plots for experiments performed on CIFAR and ImageNet datasets in experiments sections.

\renewcommand{\arraystretch}{0.0}
\begin{figure}[h]
\centering
% \vspace*{-0.4cm}
\begin{tabular}{@{}c@{}c@{}}
  \includegraphics[width=0.24\textwidth]{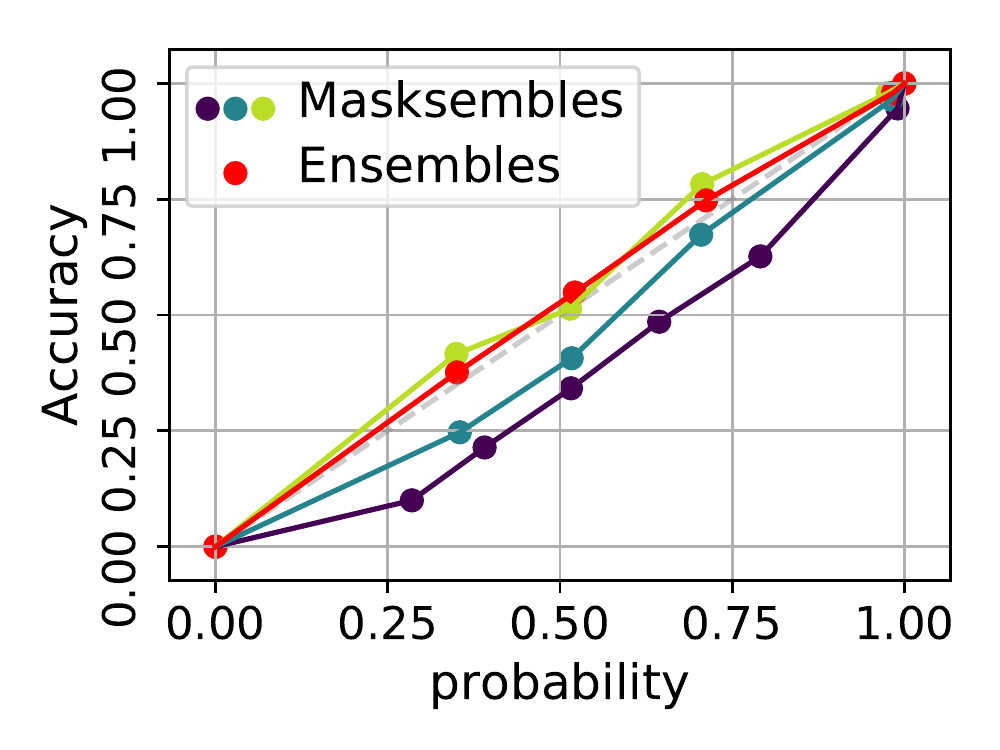} &
  \includegraphics[width=0.24\textwidth]{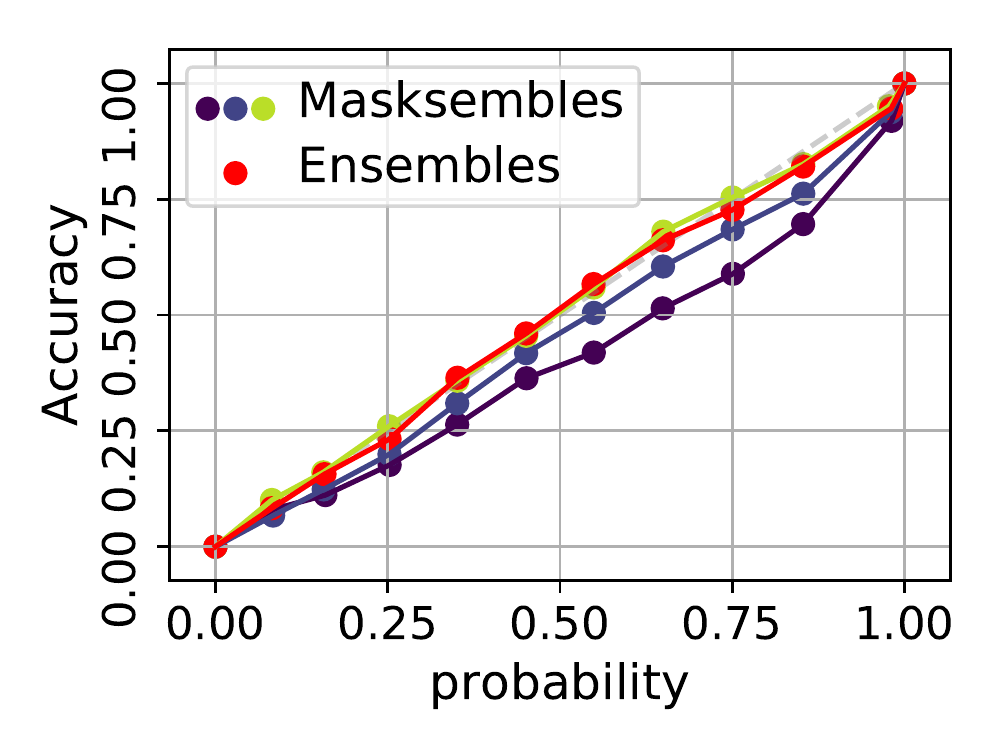} \\
\end{tabular}
\caption{{\bf Calibration plots.} Calibration results for \textbf{CIFAR10} (left) and \textbf{ImageNet} (right) test sets. Perfect calibration corresponds to $y = x$ curve. Masksembles exhibits a more ensembles-like behavior for lower values of masks overlap.}
\label{fig: toy transition}
\end{figure}
\renewcommand{\arraystretch}{1.0}
% \vspace*{-0.35cm}
%

For both datasets, the calibration plots support our claim that lower model correlation yields better calibration. Furthermore, reducing masks overlap for Masksembles enables us to match Ensembles behavior very closely.

\end{document}